\documentclass[preprint,12pt]{elsarticle}

\usepackage{amssymb}
\usepackage{amsmath}
\usepackage{booktabs}
\usepackage{multirow}
\usepackage{caption}
\usepackage{longtable} 
\usepackage{graphicx}
\usepackage{subcaption}
\usepackage{float}
\usepackage{multirow} 

\begin{document}

\begin{frontmatter}



\title{CABrain: A Correlation-Aware Brain Foundation Model for EEG Decoding}

% \author{SuGil Choi${}^\dagger$, Mooseop Kim${}^\dagger$, Chi Yoon Jeong$^{*}$} %% Author name

\author{SuGil Choi\fnref{eq}}
\author{Mooseop Kim\fnref{eq}}
\author{Chi Yoon Jeong\corref{cor1}}

\fntext[eq]{These authors contributed equally to this work.}
\cortext[cor1]{Corresponding author}

\ead{iamready@etri.re.kr}       % Chi Yoon Jeong’s email

\affiliation{organization={Electronics and Telecommunications Research Institute},%Department and Organization
            addressline={218 Gajeong-ro}, 
            city={Yuseong-gu},
            postcode={34129}, 
            state={Daejeon},
            country={KOREA}}
 
\begin{abstract}

%%%%%%%%%%%%%%%%%%%%%%
%% Max 250 words
%%%%%%%%%%%%%%%%%%%%%%

Electroencephalography (EEG) foundation models have emerged as promising paradigms for decoding brain activity through large-scale self-supervised pretraining. Notably, most existing approaches employ masked patch reconstruction, in which randomly selected EEG patches are masked and recovered by the network. However, purely random masking may limit representation learning as informative patches from highly correlated channels often remain unmasked, causing the network to underutilize weakly correlated channels or serial dependencies. Furthermore, because these models are typically based on transformer architectures, their performance is influenced by the choice of positional encoding. Although positional encoding has been extensively explored in natural language processing and computer vision, few studies have adapted it to the unique characteristics of EEG signals. To address these limitations, this study introduces a conditional positional encoding method based on graph convolutional networks (GCNs) to explicitly model inter-channel correlations, along with an adaptive masking strategy that enhances representation learning by modulating the difficulty of masked patch reconstruction. To the best of our knowledge, this is the first study to integrate graph-based representations into conditional positional encoding and introduce adaptive masking within EEG foundation model pretraining. Experimental results on ten benchmark datasets demonstrate that the proposed GCN-based positional encoding and adaptive masking strategies, both individually and in combination, achieve state-of-the-art or competitive performance, highlighting their effectiveness in EEG foundation modeling.

\end{abstract}


\end{frontmatter}

\section{Introduction}
\label{sec1}

Electroencephalography (EEG) is a fundamental sensing modality for brain--computer interfaces and healthcare applications, offering millisecond-level measurements of brain activity in a noninvasive and cost-effective manner. EEG signal analysis has been widely applied across various domains and has demonstrated substantial value. 
Representative application areas of EEG signal analysis include (1) diagnosis and monitoring of neurological disorders, such as epileptic seizure detection \cite{Boonyakitanont2020} and sleep stage classification \cite{Aboalayon2016, Jiquan2024}; (2) brain–computer interfaces based on motor imagery \cite{PhySioNet, SHU-MI} or imagined speech classification \cite{BCIC2020}; (3) cognitive and affective state assessment, including attention \cite{SEED-VIG}, stress \cite{MentalArithmetic, Sharma}, and emotion recognition \cite{SEED-V, FACED}.

Various machine learning models have been proposed for EEG signal analysis, evolving from approaches based on handcrafted features \cite{Bashashati, Yudhajit} to advanced deep learning methods, such as convolutional neural network (CNN) \cite{EEGNet, SPaRCNet, Tengfei}, graph convolutional network (GCN) \cite{Tengfei}, long short-term memory \cite{Gaowei, Hongli}, transformers \cite{Song, CNN-Transformer}, and diffusion models \cite{YilinWang}. Despite these technological advancements, learning robust and transferable EEG representations remains challenging because of (1) the low signal-to-noise ratio of EEG signals, (2) substantial inter-subject and inter-session variability, (3) heterogeneous channel configurations and reference schemes, and (4) variations in sampling rates and sequence lengths. These challenges hinder out-of-distribution generalization and cross-dataset transfer of models developed to decode brain activity. Moreover, the limited availability of large-scale annotated datasets presents an additional obstacle to achieving reliable performance.

To mitigate these challenges, recent research has increasingly explored large-scale self-supervised learning (SSL), a paradigm that has transformed fields such as natural language processing and computer vision. Using SSL, EEG encoders can learn generalizable and transferable feature representations, reducing their reliance on extensively labeled datasets. Masked EEG patch reconstruction has been widely employed as a pretext task for the self-supervised pretraining of EEG foundation models. In this technique, a subset of EEG patches is randomly masked, and the neural network is trained to reconstruct the missing patches from the remaining context. A series of representative EEG foundation models, including BENDR \cite{BENDR}, BIOT \cite{BIOT}, EEG2Rep \cite{EEG2Rep}, LaBraM \cite{LaBraM}, EEGPT \cite{EEGPT}, CBraMod \cite{CBraMod}, and EEGMamba \cite{EEGMamba}, have focused on the segmentation of EEG signals into patches, encoding of these patches, development of more efficient attention mechanisms, integration of SSL strategies to obtain generic representations from EEG signals, and incorporation of positional information of EEG patches.

\begin{figure}[t]
\centering
\includegraphics[width=1\textwidth]{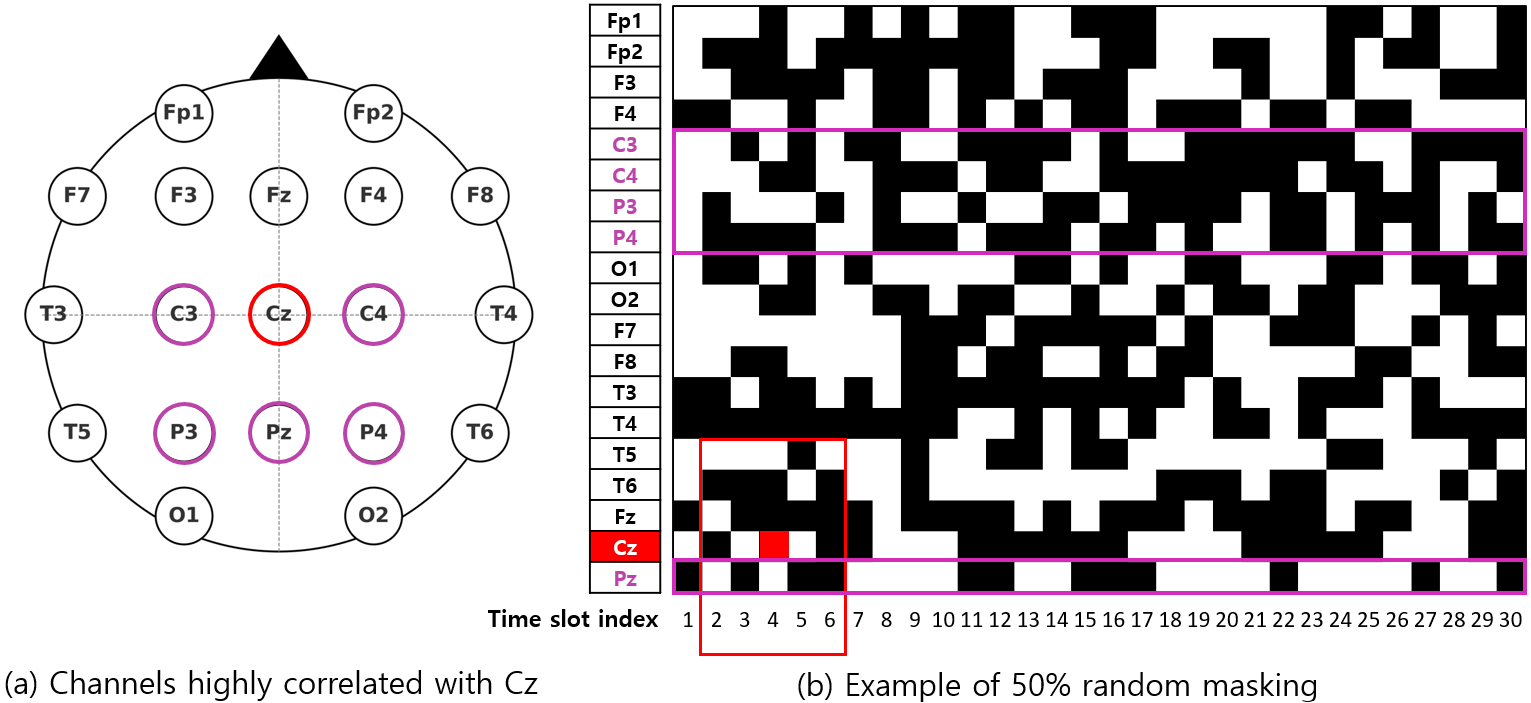}
\caption{EEG channel correlations and random masking.
(a) EEG electrode layout (10–20 system). Based on Pearson correlation coefficients estimated from the TUEG dataset, channels whose correlation coefficients are greater than or equal to 0.5 relative to Cz are highlighted by purple circles, with Cz indicated by a red circle. (b) Example of 50\% random masking used during pretraining, where masked patches are visualized in white. Rows correspond to EEG channels arranged in an arbitrary order for constructing a 2D tensor, and columns correspond to consecutive time slots. The red square indicates the receptive field of the convolution kernel used to compute the conditional positional embedding for the masked Cz patch (red cell). Purple boxes indicate patches belonging to channels strongly correlated with Cz. This example illustrates that many channels strongly correlated with Cz may lie beyond the receptive field of the convolutional kernel and are unlikely to be masked simultaneously under random masking.}\label{fig1}
\end{figure}

Among the aforementioned frameworks, CBraMod \cite{CBraMod} is notable for its evaluation on more than ten heterogeneous test datasets, where it achieved state-of-the-art performance at the time of publication. One of the main contributions of CBraMod is the development of asymmetric conditional positional encoding (ACPE). Conditional positional encoding (CPE) \cite{CPE} was originally proposed for vision transformers. Unlike conventional fixed positional encodings, CPE is dynamically computed from the input data, enabling greater adaptability across diverse datasets. In CBraMod, the raw EEG signal from each channel is segmented into small chunks of predetermined duration, and each chunk is defined as an EEG patch. A two-dimensional (2D) grid is constructed by stacking patches from different channels in an arbitrary order, with each patch representing a single element. ACPE is applied to this grid representation, enabling convolutional operations to treat EEG patches analogously to image patches.

A key limitation of ACPE is that it treats the constructed grid representation as an image and applies convolution. While pixels within local neighborhoods are strongly correlated owing to spatial continuity in images, EEG grid representations do not necessarily ensure strong correlations between adjacent patches. Figure 1 illustrates this limitation. Specifically, Figure 1(a) highlights the channels strongly correlated with Cz, whereas Figure 1(b) shows that most of these correlated channels lie outside the receptive field of the convolutional kernel in the stacked 2D tensor representation. The channel correlations are estimated using the Temple University Hospital EEG Corpus (TUEG) \cite{TUEG} dataset, which is also used as the training dataset in this study. Consequently, convolution may aggregate information from weakly correlated channels while overlooking more informative but spatially distant channels in the stacked 2D tensor representation. Thus, the positional embeddings may not accurately reflect the underlying inter-channel relationships, reducing their effectiveness as positional cues for EEG representation learning. In contrast, graph convolution operates on neighborhoods defined by inter-channel correlations rather than arbitrary spatial proximity, enabling positional embeddings to be generated from more meaningful local relationships.

Motivated by this observation, we propose a novel GCN-based CPE approach. Additionally, we examine the effectiveness of SSL approaches based on masked patch reconstruction in capturing the underlying structures of EEG signals. The conventional reconstruction of randomly masked patches involves hiding certain patches at random and training the model to recover them using the surrounding context. However, random masking inherently limits representation learning because informative signals, such as patches from strongly correlated channels, may remain unmasked. In the random masking example shown in Figure 1(b), patches from channels strongly correlated with Cz are not masked simultaneously within the same time slot. Even in the most challenging case (time slot 1), one strongly correlated channel (Pz) remains observable. Across the 30 time slots shown in the example, an average of 2.8 strongly correlated channels remain unmasked per time slot. Consequently, highly informative correlated-channel signals remain available for reconstructing masked Cz patches. To enhance the representation learning capability of neural networks, information from weakly correlated channels and serial correlations must be considered. However, the consistent availability of information from strongly correlated channels may prompt neural networks to underutilize weakly correlated channels or serial correlations during pretraining. To address this, we introduce a novel adaptive masking strategy for SSL that adjusts the difficulty of masked patch reconstruction, encouraging the utilization of information from weakly correlated channels and serial correlations.

Both proposed methods, i.e., GCN-based CPE and adaptive masking, leverage the correlation information inherent in EEG signals. Unlike absolute measures like the number or arrangement of EEG electrodes and signal amplitude, correlation is less affected by dataset variability. The key contributions of this work can be summarized as follows.
\begin{itemize}
    \item \textbf{GCN-based CPE.} We introduce a CPE method based on a GCN, in which graph edges are determined by channel correlations. By performing convolution exclusively on strongly correlated channels, the proposed positional encoding effectively integrates local correlation, capturing more invariant features and improving adaptability to unseen EEG electrode configurations during foundation model pretraining.
    \item \textbf{Adaptive masking strategy for effective representation learning.} We propose a novel adaptive masking strategy for SSL that affords control over the reconstruction difficulty of masked patches and potentially facilitates a deeper understanding of EEG signals. While adaptive masking has been extensively studied in large language models (LLMs) and computer vision, to the best of our knowledge, this study represents the first attempt at applying adaptive masking for self-supervised pretraining of EEG foundation models.
    \item \textbf{Comprehensive experiments on diverse datasets.} Each proposed approach achieves performance comparable or superior to that of recent state-of-the-art frameworks. The combination of both methods results in performance superior to that achieved by either approach individually.
\end{itemize}

\section{Related work}
\label{sec2}

\subsection{EEG Foundation Model}
\label{subsec2-1}

EEG foundation models differ in several aspects, including backbone architecture and pretraining objective. From an architectural perspective, transformer-based models have been widely adopted, although recent studies have also explored state-space models as an alternative backbone architecture. However, architectural differences alone do not fully explain the methodological diversity of recent EEG foundation models, as models with similar architectures often adopt different self-supervised learning objectives. In contrast, pretraining objectives determine what information the model is encouraged to learn from unlabeled EEG signals and have thus evolved along several directions. Therefore, the following review categorizes EEG foundation models according to their pretraining objectives.

Based on pretraining objectives, existing EEG foundation models can be broadly classified into three groups: contrastive learning, autoregressive prediction, and signal reconstruction-based pretraining. Contrastive learning methods aim to learn invariant EEG representations by maximizing the agreement between related samples while separating unrelated samples. Representative examples include BENDR \cite{BENDR} and BIOT \cite{BIOT}. Autoregressive prediction follows a causal next-token prediction paradigm, wherein future EEG tokens are predicted from preceding context using causal masking, similar to the training of LLMs. Representative examples include Neuro-GPT \cite{Neuro-GPT}, NeuroLM \cite{NeuroLM}, BrainGPT \cite{BrainGPT}, and THD-BAR \cite{THD-BAR}. Among these three groups, signal reconstruction-based pretraining has emerged as the dominant paradigm. In signal reconstruction-based pretraining, a portion of the input is masked and the model is trained to recover the missing information from the remaining context. Depending on the reconstruction target, existing approaches can be further divided into three categories: raw-signal reconstruction, representation reconstruction, and frequency-domain reconstruction.

The first category reconstructs the raw EEG signal. Models such as FoME \cite{FoME}, CBraMod \cite{CBraMod}, EEGMamba \cite{EEGMamba}, CSBrain \cite{CSBrain}, REVE \cite{REVE}, and LUNA \cite{LUNA} learn to recover masked EEG segments using reconstruction losses applied to the original signal. Unlike representation-based objectives, raw-signal reconstruction operates  on the original EEG measurements and encourages the preservation of information contained in the input signal.

\begin{figure}[t]
\centering
\includegraphics[width=1\textwidth]{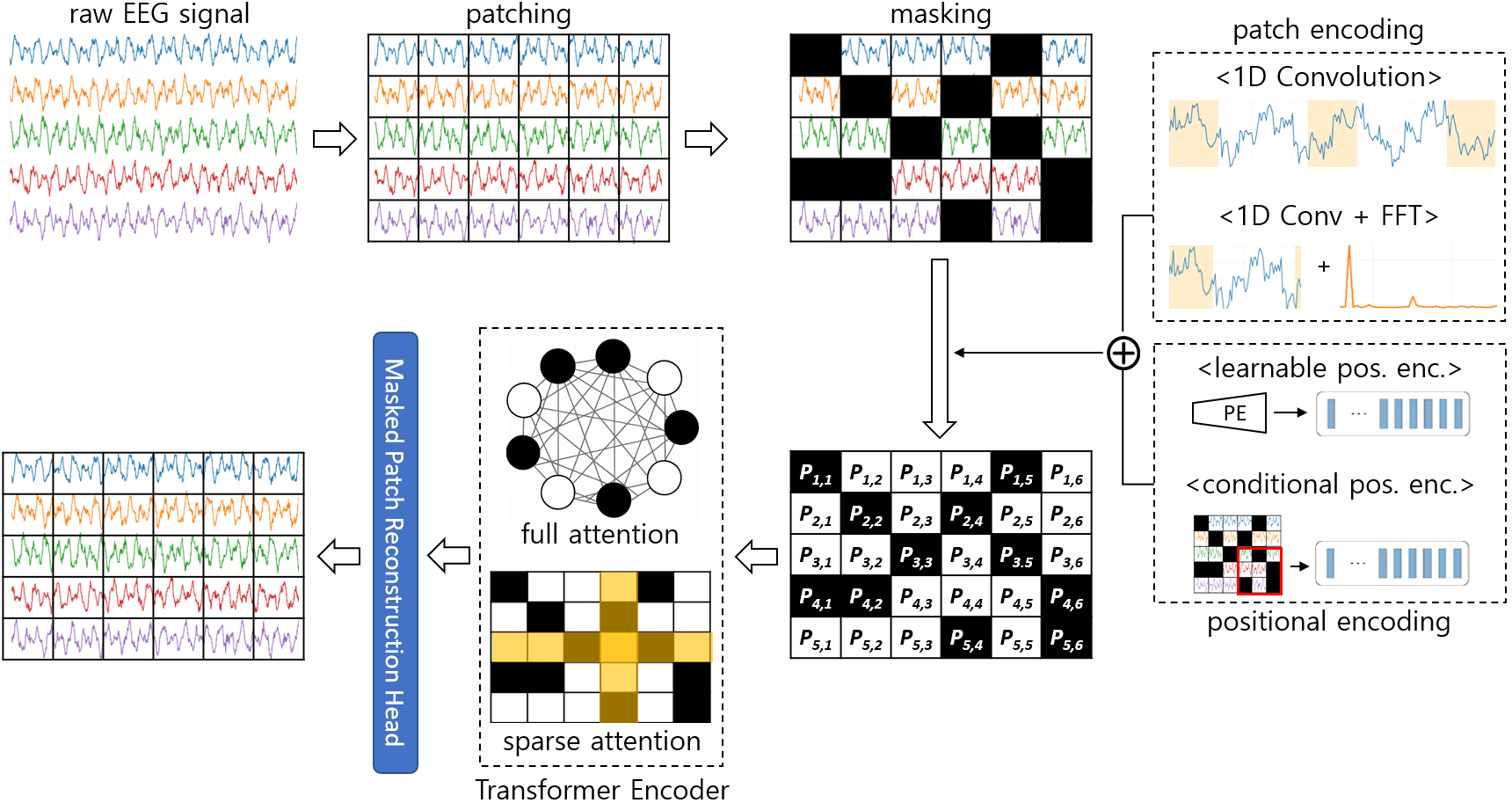}
\caption{General framework of raw-signal reconstruction-based EEG foundation models.}\label{fig2}
\end{figure}

The second category reconstructs intermediate EEG representations rather than the original signal itself. Depending on the representation space, the reconstruction target may consist of continuous latent embeddings or discrete codebook representations. In embedding reconstruction, EEG patches are transformed into latent representations using a representation encoder, and the model is trained to predict the masked latent features. Representative models include EEGPT \cite{EEGPT}, EEG2Rep \cite{EEG2Rep}, and MIRepNet \cite{MIRepNet}. In contrast, codebook-based approaches first map EEG representations into discrete codebook entries through vector quantization and then predict the corresponding codebook indices. Representative examples include LaBraM \cite{LaBraM}, BrainOmni \cite{BrainOmni}, EpilepsyFM \cite{EpilepsyFM}, and CodeBrain \cite{CodeBrain}. By defining the reconstruction target in a learned representation space rather than the original signal domain, these approaches encourage the model to focus on higher-level EEG characteristics while reducing the influence of low-level signal variability.

The third category focuses on reconstructing frequency-domain information, where the reconstruction target is formulated using spectral representations instead of time-domain signals or learned representations. BrainBERT \cite{BrainBERT}, EEGFormer \cite{EEGFormer}, BrainWave \cite{BrainWave}, and Uni-NTFM \cite{Uni-NTFM} are representative examples. Because spectral characteristics play an important role in many EEG decoding tasks, frequency-domain objectives encourage the model to capture oscillatory patterns that may not be readily observable in the time domain.

Although reconstruction targets differ, many EEG foundation models adopt a broadly similar pretraining pipeline, as illustrated in Figure 2. Continuous EEG recordings are first segmented into fixed-length temporal patches for each channel. These patches are then transformed into feature representations using temporal, spectral, or hybrid encoders. After positional information is incorporated, the resulting patch embeddings are processed by a backbone network, typically a transformer encoder. The model is then trained to infer masked patches from the surrounding context according to the chosen reconstruction target.

\subsection{Positional Encoding}
\label{subsec2-2}

In natural language processing and computer vision, positional encoding plays a critical role in enabling transformer-based models to capture the order and spatial structure of the input data. The most commonly used approach is absolute positional encoding, in which fixed or learnable vectors are assigned to specific positions within an input sequence or image grid. This strategy, first popularized in natural language processing, allows the model to differentiate between tokens and patches based on their absolute locations. In contrast, relative positional encoding represents the distance or offset between tokens or patches, enabling translational invariance.

Recently, CPE \cite{CPE} has been proposed as a positional encoding scheme in which encodings are dynamically generated based on the local neighborhood of input tokens rather than being fixed or predefined. CPE is typically implemented using a positional encoding generator, often using lightweight 2D convolutions, making it efficient and easy to integrate into transformer architectures. In contrast to absolute or relative encodings, CPE can naturally generalize to input sequences that are longer than those observed during training. Originally introduced for vision tasks, CPE was designed to satisfy the following three desirable properties:
\begin{itemize}
\item exhibit permutation variance while preserving translation equivariance
\item generalize to sequences longer than those encountered during training
\item retain a certain degree of absolute positional information to ensure strong performance
\end{itemize}
Furthermore, Saijo and Ogawa \cite{Kohei} referred to CPE as ``no positional encoding'' and reported that it achieved favorable performance on speech separation datasets involving sequences longer than those used during pretraining.

\subsection{Masking Strategies in SSL}
\label{subsec2-3}
In natural language processing, SSL has been widely implemented using masked language modeling (MLM). Motivated by this approach, masked image modeling (MIM) has been extensively explored in computer vision. MLM predicts missing words in a text sequence based on the surrounding context, whereas MIM reconstructs the masked regions in an image by leveraging the visible information. Although random masking has been predominantly used in both MLM \cite{BERT, MASS} and MIM \cite{BEIT, KaimingHe, Simmim}, adaptive masking has received increasing attention owing to its potential to improve the representation learning capacity.

Recent work on MLM \cite{APOLLO} has shown that not all masked tokens require the same degree of reasoning for prediction, and that logical words, such as ``more'', ``and'', and ``hence'' entail more complex reasoning than ordinary words. Experiments demonstrated that preferentially masking logical words can foster more effective reasoning than random masking. Kohli et al. \cite{Vanshpreet} proposed a domain-specific masking approach in which tokens associated with specific concepts are selectively masked to enable continual learning with reduced training overhead.

Among recent MIM research, Feng and Zhang \cite{ZhanzhouFeng} demonstrated that fixed masking patterns constrain the modeling of visual cues. Therefore, the authors proposed an evolved part-based masking method using an adaptive part-partition module that enables more general visual cue modeling. In the work of Bandara et al. \cite{AdaMAE}, a categorical distribution was estimated over space--time-patch tokens and those expected to yield larger reconstruction errors were rewarded and chosen as visible tokens. Shah et al. \cite{CSMAE} introduced a pretraining approach for cataract surgery video analysis, where token selection and masking were guided by spatiotemporal importance instead of random choice.

Dai et al. \cite{EEG2Rep} argued that the absence of structure in randomly masked data hinders the learning of semantic representations. The authors proposed the semantic subsequence preserving method, which focuses on selecting regions to preserve rather than regions to mask. However, as the selection of preserved regions is random, the method remains closer to random masking than adaptive masking approaches used in LLMs and computer vision.

More recently, Tang et al. \cite{EpilepsyFM} proposed a brain-region masking strategy in which groups of channels corresponding to predefined anatomical brain regions are masked. By removing spatially related channels together, the model is encouraged to infer missing information from inter-regional dependencies rather than relying on isolated channels. Unlike random masking, this approach explicitly incorporates domain knowledge regarding brain organization into the masking process.

\section{Method}
\label{sec3}

This section describes the proposed framework, CABrain, incorporating novel positional encoding and mask generation methods within the pretraining framework of an EEG foundation model. First, a brief overview of CABrain is provided, and Subsections \ref{subsec3-1} and \ref{subsec3-2} outline the positional encoding and masking strategies, respectively.

The EEG signal from each channel is segmented into one-second intervals. Assuming a sampling frequency of 200 Hz, each one-second patch consists of 200 data points. A predetermined percentage of patches is masked, and the corresponding feature vectors are replaced with zeros. Unlike prior studies that predominantly relied on random masking, the proposed approach adopts an adaptive mask generation strategy designed to enhance the representation learning capability. After masking, patch encoding proceeds through two pathways: A temporal pathway applies one-dimensional CNNs to extract time-domain features, whereas a spectral pathway applies a fast Fourier transform for frequency-domain feature extraction. The two types of features are merged via element-wise summation. Positional embeddings computed using the proposed method are added to yield a patch embedding for the transformer encoder. Let $C$ denote the number of EEG channels and $N$ be the number of patches per channel. Each EEG patch is denoted by $x \in \mathbb{R}^{\ell}$, and the set of patches is $X = \{\, x_{i,j} \mid i \in [1,2,\ldots, C],\; j \in [1,2,\ldots, N] \}$. The EEG patches are first processed by the transformer encoder through self-attention, and subsequently passed to a masked EEG patch reconstruction head. To reduce the computational cost pertaining to full attention, the model adopts crisscross attention, a sparse attention variant originally introduced in CBraMod \cite{CBraMod}.

\subsection{GCN-Based CPE}
\label{subsec3-1}

CPE \cite{CPE} uses CNNs to generate positional embeddings conditioned within a local spatiotemporal context. Based on this concept, CBraMod \cite{CBraMod} introduced ACPE, in which per-channel EEG signals are stacked into a 2D matrix (with arbitrary channel ordering) and convolved using an asymmetric 19×7 kernel. The patch grid is zero-padded by nine elements on the top and bottom, and by three elements on the left and right. Figure 3 illustrates the convolutional filter used to compute the positional embedding for the third patch of the Cz channel. Channels whose correlation with Cz is $\geq 0.5$ are highlighted in purple. Although CBraMod with ACPE achieves state-of-the-art results for a range of downstream tasks, it exhibits two limitations.

\begin{figure}[t]
\centering
\includegraphics[width=1\textwidth]{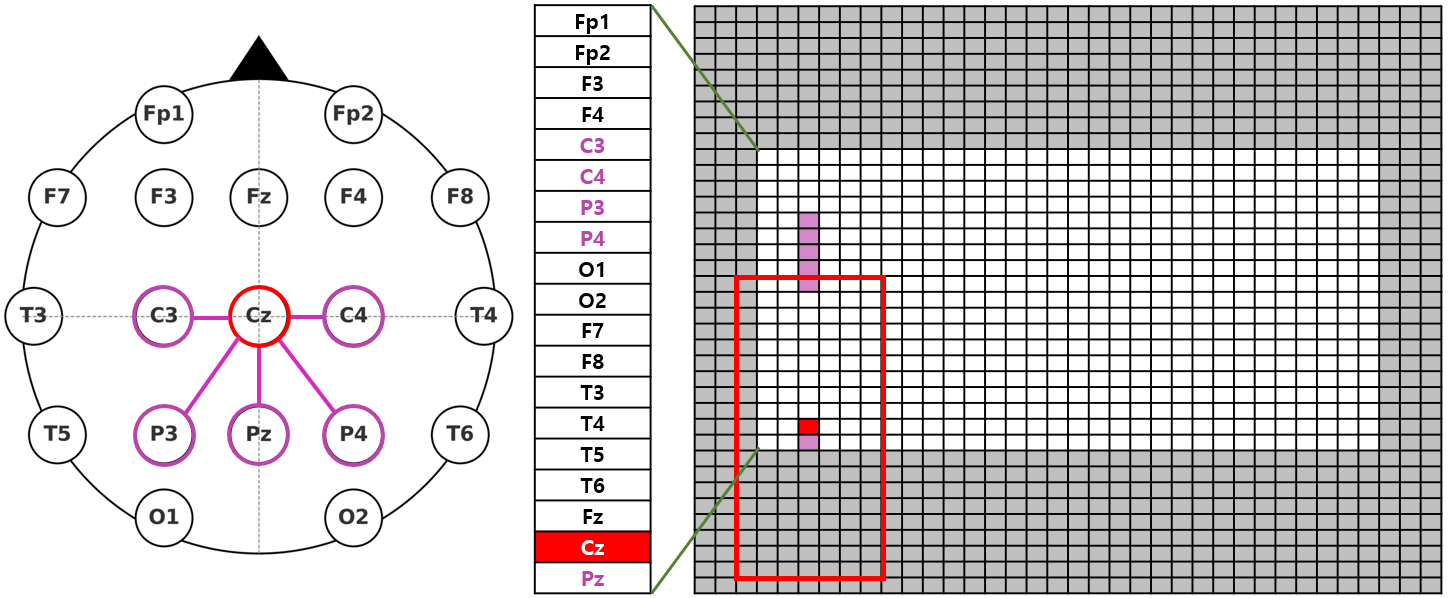}
\caption{Application of a 19×7 convolution to a 2D tensor formed by stacking EEG channels in an arbitrary sequence.}\label{fig3}
\end{figure}

First, convolution on a 2D tensor obtained from arbitrary channel stacking may not reliably capture informative local structures. Positional encoding is essential in transformer architectures because self-attention alone is permutation-invariant and cannot distinguish the positions of input tokens. Therefore, positional embeddings should provide location-dependent cues that enable the model to differentiate one token from another while preserving meaningful structural relationships. CPE \cite{CPE} was originally designed for images, which typically exhibit strong local pixel correlations. The correlation coefficient analysis in \cite{LongxuanDai} indicated that adjacent pixels in images are sufficiently strongly correlated such that the values of neighboring pixels can be largely inferred from a single pixel. Because positional embeddings are computed by convolving neighboring pixels, these strong local correlations allow the convolution operation to produce stable and consistent positional cues that reflect the underlying spatial structure of the image. However, as shown in Figure 3, convolution over the 2D tensor formed by arbitrarily stacking EEG channels leverages only two channels whose correlation with Cz is $\geq 0.5$. The remaining lower-correlation channels may provide little informative signals, potentially introducing noisy information. Consequently, the convolution may generate inconsistent positional embeddings that do not capture the underlying correlation structure of EEG channels, limiting their effectiveness as positional cues for transformer-based EEG representation learning.

Second, zero padding along the boundaries may leak nontrivial cues about the absolute position rather than only weak signals. Among the three requirements identified in \cite{CPE} for effective positional encoding in vision tasks, the third is the capacity to slightly convey the absolute positions. In conventional vision tasks, boundary padding occupies only a small fraction of the receptive field, such that convolution outputs are still dominated by the underlying image content and padding provides only weak positional cues. In contrast, the ACPE employed in CBraMod pads the EEG grid by nine elements vertically and three elements horizontally for a 19×7 convolution. Thus, a large portion of the receptive field for boundary patches may consist of padded zeros. In this case, the convolution output may be influenced more strongly by the padding pattern than by the EEG patches themselves, causing positional embeddings to encode excessive absolute positional information rather than only weak boundary cues, deviating from the original design objective of CPE.

To address the first issue, in this study, a graph is constructed using EEG electrodes as nodes, and a GCN-based CPE strategy is adopted. To address the second issue, we reconsider whether absolute positional information is necessary for EEG representation learning. Among the requirements for effective CPE \cite{CPE}, the condition that positional encodings provide absolute position ``to a certain degree'' appears to be motivated primarily by empirical observations, with limited theoretical justification. While positional encodings in natural language processing represent token locations within a sequence, EEG electrodes represented as graph nodes are characterized by their topological relationships rather than by a well-defined absolute position. Consequently, introducing absolute positional cues may offer limited gains in EEG graphs and may even promote overfitting to dataset-specific electrode layouts. Therefore, the proposed GCN-based CPE is designed without explicit absolute positional information, relying solely on structural relationships among electrodes. This design preserves generalization performance while avoiding the excessive positional bias introduced by zero padding.

Representing EEG data as a graph for GCNs requires the definition of two components: node features and an adjacency matrix. EEG channels are defined as nodes, whereas the patch embedding serves as a node feature. For datasets with different channel configurations, graphs are reconstructed using the channels available in each dataset rather than being mapped to a common channel subset. Despite differences in the numbers of nodes and edge connections, the pretrained GCN weights remain applicable because graph convolution is defined through shared neighborhood aggregation and weight matrices that are independent of the specific graph topology. As GCN parameters operate on feature dimensions rather than individual node identities, the same pretrained weights can be directly applied to graphs with different numbers of nodes (e.g., 62 or 6 channels) and connectivity patterns.

For an input represented as a $C$ × $N$ matrix, $N$ graphs are constructed to compute the positional embeddings for each time slot. Notably, an adjacency matrix encoding inter-channel relations can be constructed using either distance-based or correlation-based methods. The distance-based method exploits the physical geometry of EEG electrodes. A method for constructing a distance-based adjacency matrix can be found in [60]. However, distance-based methods have two limitations. First, they can only be applied when feature values are generated by encoding signals from individual channels and cannot be readily employed when bipolar montage channels are used. Because bipolar montage channels represent voltage differences between electrode pairs, a meaningful physical distance between bipolar channels cannot be defined. Second, the physical locations of electrodes may vary across individuals and recording systems owing to differences in head size, electrode placement, and acquisition protocols. Moreover, even when subjects perform the same task, the spatial distribution of EEG responses may vary across individuals, resulting in different sets of channels carrying the most informative signals. Consequently, physically adjacent electrodes do not necessarily exhibit similar functional roles across individuals. Such inter-subject variability can lead to inconsistencies in distance-based graph structures and may reduce the robustness and generalization capability of positional encodings across heterogeneous datasets.

Considering these aspects, this study adopts a correlation-based method for constructing an adjacency matrix. Unlike distance-based graph construction methods that rely on electrode coordinates, the proposed approach constructs graphs solely from inter-channel correlation statistics. Therefore, the graph structure is independent of the physical arrangement and total number of electrodes, making it applicable to datasets with heterogeneous channel configurations. The absolute value of the Pearson correlation coefficient between node features was used to determine graph connectivity. The Pearson correlation coefficients are estimated from the TUEG \cite{TUEG} dataset during foundation model pretraining. A separate set of coefficients is estimated using only the training split of each target dataset during downstream fine-tuning. In general, graph connectivity can be determined using various strategies, including threshold-based edge selection and top-K neighbor selection. In this study, we adopt the latter approach, where each node is connected to its K most strongly correlated neighbors together with the self-connections. Various prior studies have similarly defined graph structures using correlations between nodes \cite{SiyiTang, JinpeiHan}. Figure 4 illustrates a simplified case in which nodes are connected to the three most strongly correlated nodes, and positional encodings are subsequently computed using a GCN.

\begin{figure}[t]
\centering
\includegraphics[width=1\textwidth]{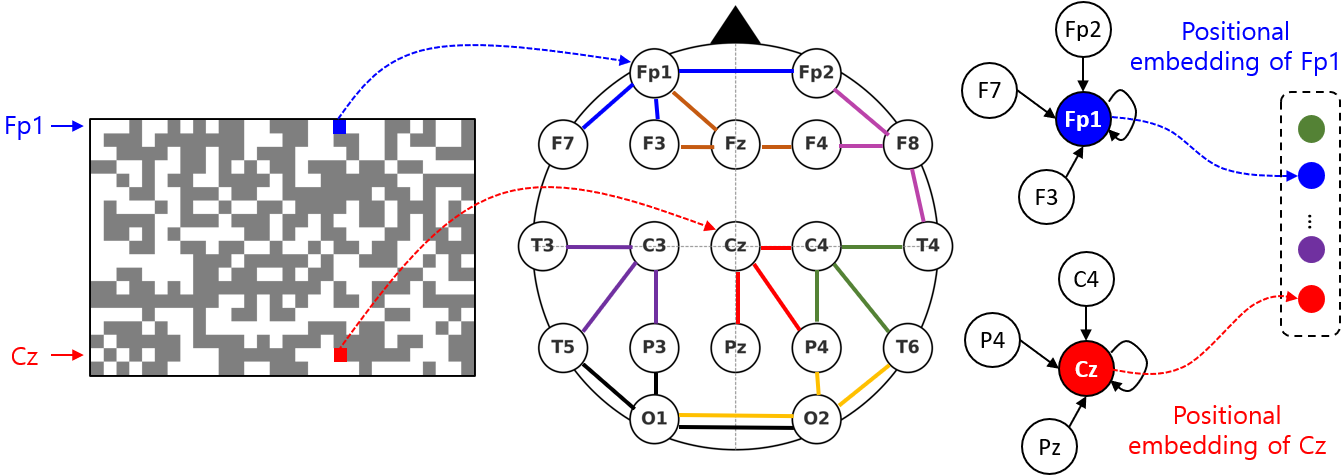}
\caption{Graph convolution with edges defined by the top three inter-channel correlations.}\label{fig4}
\end{figure}

When computing positional encodings with a GCN, the patch embeddings to be employed as node features must be obtained after the masking step. If the patch embeddings are computed from the data before masking and fed into the GCN, the resulting positional embeddings may introduce a unique challenge. When these embeddings are added to the masked patch embeddings and passed through the transformer encoder for masked patch reconstruction pretraining, the GCN may encode information tailored to the reconstruction objective rather than providing meaningful positional cues. In extreme cases, the GCN may degenerate to reproduce its input patch embeddings. The transformer encoder and reconstruction head then minimize reconstruction loss by overfitting the training data while failing to learn useful positional representations. In contrast, when zero-filled patch embeddings derived from masking are included as node features, making the GCN preserve as much of the input information as possible does not benefit the reconstruction task. Instead, the presence of zero-filled embeddings implicitly induces an edge-drop effect in GCN computation. Although the edges are not explicitly removed, multiplication with zero-filled features effectively cancels certain connections. As reported by Rong et al. \cite{EdgeDrop}, randomly dropping edges during GCN computation helps mitigate overfitting and over-smoothing. Incorporating zero-filled embeddings may thus provide a similar regularization effect.

Based on the above discussion, the proposed GCN-based CPE can be formulated as follows:

\begin{equation}
\mathbf{x}_{v}^{(l+1)}
=
\mathbf{\Theta}_{l}^{\top}
\sum_{u \in \mathcal{N}_{\mathrm{corrTop}K}(v)\cup\{v\}}
\frac{\widetilde{A}_{vu}}
{\sqrt{\widetilde{d}_{v}\widetilde{d}_{u}}}
\mathbf{x}_{u}^{(l)}.
\end{equation}

Considering an input represented as a \(C \times N\) matrix, \(N\) graphs
were constructed to compute the positional embeddings for each time slot.
However, for simplicity, we assume a single graph comprising \(C\) nodes. Here, \(\mathbf{x}_{u}^{(l)}\) denotes the feature of node \(u\) at layer \(l\),
\(\mathcal{N}_{\mathrm{corrTop}K}(v)\) represents the set of top-\(K\) nodes
that are most strongly correlated with node \(v\),
and \(\mathbf{\Theta}_{l}\) is the trainable weight matrix of the \(l\)-th
graph convolution layer.

Let \(A \in \{0,1\}^{C \times C}\) denote the binary adjacency matrix
constructed from the correlation-based top-\(K\) graph, where
\(A_{vu}=1\) if \(u \in \mathcal{N}_{\mathrm{corrTop}K}(v)\) and
\(A_{vu}=0\) otherwise.
The self-loop augmented adjacency matrix is defined as
\(\widetilde{A}=A+I_C\), and
\(\widetilde{d}_v=\sum_{u=1}^{C}\widetilde{A}_{vu}\)
denotes the degree of node \(v\) after adding self-connections.
Pearson correlation coefficients were used only to determine
\(\mathcal{N}_{\mathrm{corrTop}K}(v)\) during graph construction.
No Pearson correlation coefficient was used as an edge weight; all retained
edges were treated as binary connections, i.e.,
\(\widetilde{A}_{vu}\in\{0,1\}\).
Accordingly, the graph convolution performs symmetric degree-normalized
aggregation over the unweighted top-\(K\) correlation-based neighborhood.

\subsection{Adaptive Masking Strategy in SSL}
\label{subsec3-2}

\begin{figure}[H]
\centering
\includegraphics[width=1\textwidth]{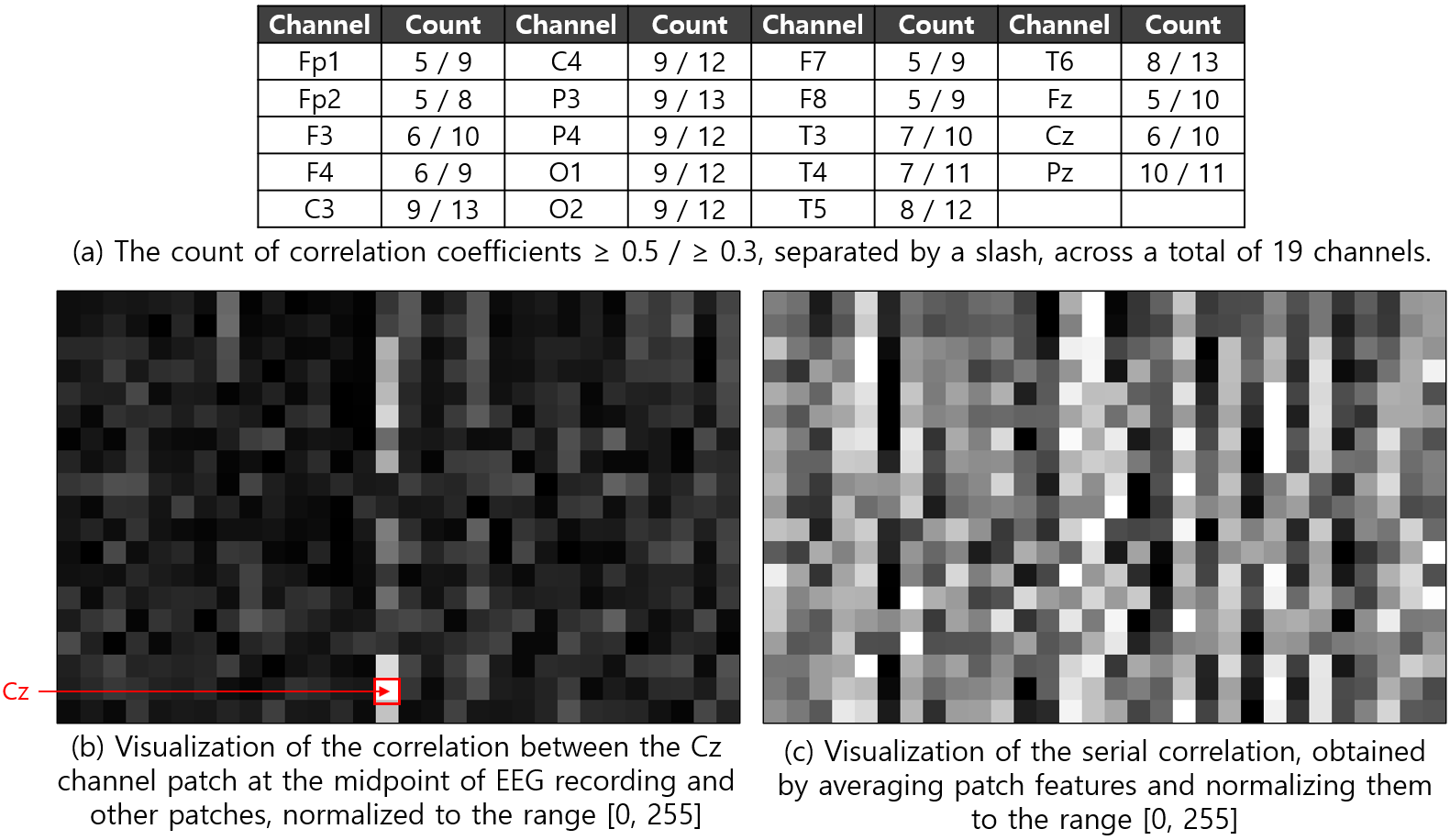}
\caption{Analysis of channel and serial correlation in EEG signals.}\label{fig5}
\end{figure}

The selection of patches for masking is critical for learning transferable representations from unlabeled EEG signals. Robust and generic EEG representations can be learned from reconstruction tasks of appropriate difficulty. To design such a masking method, the information available for inferring the original data is first analyzed. Although the inference process of AI models cannot be fully determined, an approximate understanding can guide the development of masking strategies that promote more effective reasoning than random masking.

To this end, both the inter-channel and serial correlations of EEG signals are exploited. Inter-channel correlation indicates the relationship between EEG channels, whereas serial correlation defines the temporal correlation within a single channel over time. As shown in Figure 5(a), for each of the 19 channels, the proportions of Pearson correlation coefficients that are greater than or equal to 0.5 and 0.3 with other channels are 37.95\% and 56.79\%, respectively. In other words, various channels exhibit nonnegligible correlations, making it unlikely that all of them are masked simultaneously during random masking. Hence, the information from moderately or strongly correlated channels can likely be used to reconstruct randomly masked patches.

Figure 5(b) shows that the channel correlations are stronger among the simultaneously recorded signals, whereas those across different time slots remain weak. This suggests that correlations of channels from different time slots are less likely to be leveraged for masked patch reconstruction. In Figure 5(c), patch colors represent the average values of the elements in the feature vector, with similar colors across consecutive patches suggesting a strong serial correlation. Although horizontally adjacent patches often exhibit similar colors, abrupt changes occur frequently. Therefore, relying on serial correlation may occasionally increase the reconstruction error, and when strongly correlated channels are available, their contribution to pretraining may be limited.

In summary, Figure 5 reveals two key characteristics of EEG patch reconstruction. First, strongly or moderately correlated channels are likely to remain available under random masking, causing the model to rely on these highly informative channels while underutilizing weakly correlated channels. Second, serial correlations provide useful temporal cues, albeit with reduced utility and reliability compared with strongly correlated channels. Consequently, random masking may not adequately encourage the model to simultaneously exploit weakly correlated channels and temporal dependencies. To promote more robust representation learning, the masking strategy should intentionally reduce the availability of strongly correlated channel information while preserving sufficient temporal context for reconstruction. Therefore, the proposed masking strategy aims to create reconstruction scenarios in which weakly correlated channels and temporal dependencies become necessary sources of information rather than optional alternatives.

The proposed kernel density estimation (KDE)-based masking strategy is designed to achieve two objectives.
\begin{enumerate}
\item Increasing the probability of selecting already heavily masked time slots. Even when a time slot already contains many masked patches, the proposed KDE scheme can further increase its sampling probability if neighboring time slots still contain sufficient unmasked patches. In this scenario, certain time slots become heavily masked while nearby time slots remain observable. This reduces the availability of strongly correlated channel information within the heavily masked time slots and encourages the model to rely more on weakly correlated channels for reconstruction.
\item Preventing simultaneous heavy masking of adjacent time slots. If multiple adjacent time slots become heavily masked simultaneously, temporal dependencies can no longer be exploited for reconstruction. The KDE therefore decreases the probability of selecting time slots surrounded by heavily masked neighbors. This mechanism limits long runs of consecutive masking and preserves temporal context, increasing the likelihood that both weakly correlated channels and serial dependencies are leveraged during reconstruction.
\end{enumerate}

Together, these objectives are intended to create reconstruction scenarios in which strongly correlated channel information becomes insufficient while temporal context remains available. Such scenarios encourage the model to simultaneously exploit weakly correlated channels and serial dependencies, which are rarely required under conventional random masking. The proposed KDE facilitates the generation of these scenarios by incorporating the distribution of unmasked patches across neighboring time slots into the masking process.

\begin{figure}[t]
\centering
\includegraphics[width=1\textwidth]{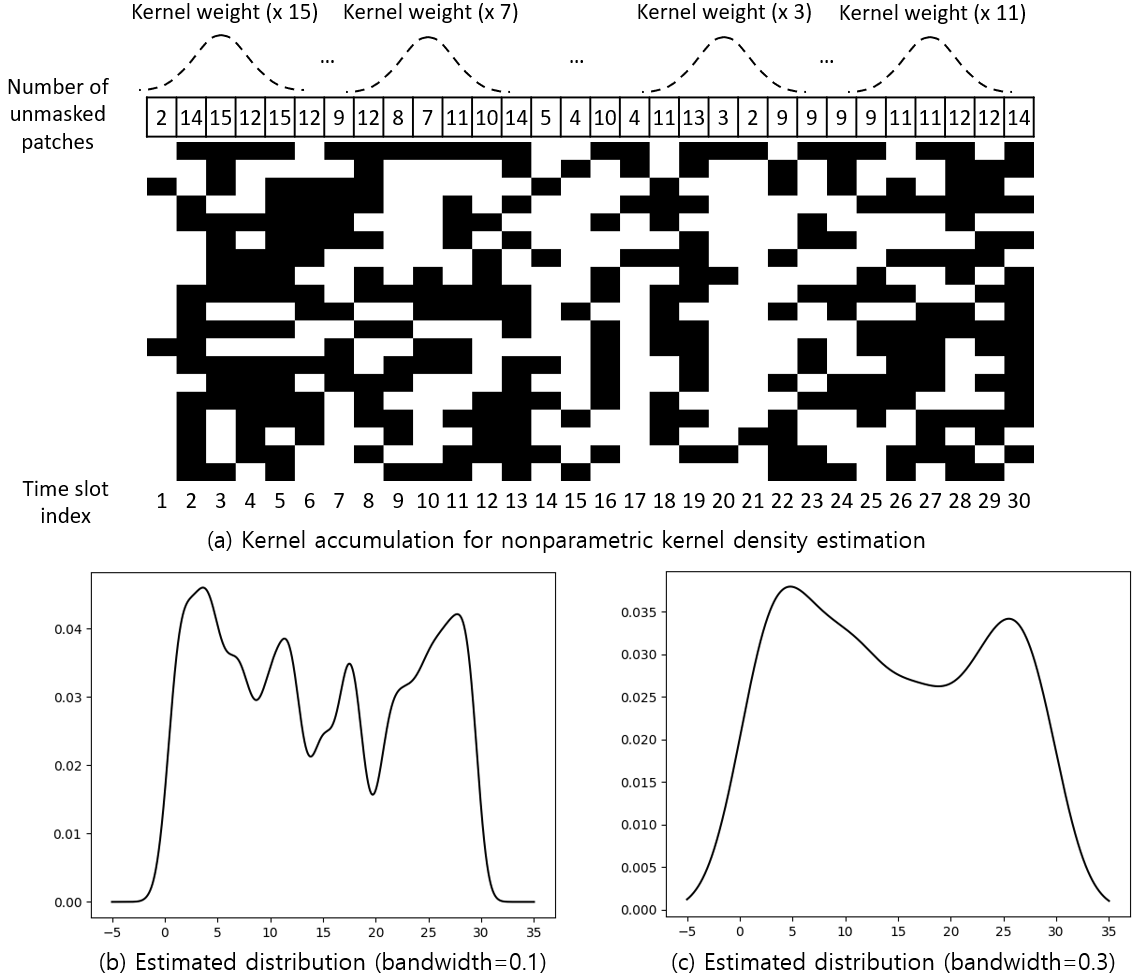}
\caption{Estimation of nonparametric kernel density for sampling time slots as part of adaptive mask generation.}\label{fig6}
\end{figure}

The proposed method first performs KDE by placing the kernels in proportion to the number of unmasked patches within each time slot. Based on the resulting probability distribution, a time slot is sampled, and one unmasked patch is randomly chosen from that time slot. This density estimation and sampling procedure is iteratively performed until the desired number of masked patches is obtained. In Figure 6(a), black patches indicate unmasked regions, and numbers above the grid represent the number of unmasked patches per time slot. The notation ``× K'' indicates that the kernel centered at a specific time slot is accumulated K times when estimating the density. Therefore, time slots containing more unmasked patches contribute more strongly to the estimated distribution and also increase the probability of selecting neighboring time slots. When 50\% of the patches are randomly masked, it is highly unlikely that a single time slot will contain 80–90\% of the masked patches. This scenario occurs in the 20th and 21st slots, as shown in Figure 6(a). Replacing random sampling with a KDE-based proposal distribution, $\hat{f}_h(t)$, which incorporates neighboring time slots, increases the density around the 20th and 21st slots. This allows masking to occur at proportions well above the average in these slots. Moreover, this increase in sampling probability occurs only when neighboring time slots have sufficient unmasked patches. This configuration enhances the likelihood of exploiting both serial correlations and weakly correlated channels for reconstruction.

The density shape depends on the kernel bandwidth, as shown in Figures 6(b) and 6(c) for bandwidths of 0.1 and 0.3, respectively. When the kernel bandwidth is extremely small, the nonparametric approach converges to random masking. However, with increasing bandwidths, the probability of selecting a time slot increases, even if it already contains many masked patches, provided that its neighboring slots have more unmasked patches. Therefore, appropriate bandwidth selection is crucial for generating masks that align with the intended objective of leveraging both serial dependencies and weakly correlated channel information.

Formally, the nonparametric kernel density estimator is defined as
\begin{equation}
\hat{f}_h(t) = \frac{1}{Z h} \sum_{i=1}^C \sum_{j=1}^N 
\mathbb{I}_{\{x_{ij} \notin Masked\}}
K\!\left(\frac{t - j}{h}\right),
\end{equation}
where $C$ denotes the number of EEG channels, $N$ represents the number of time slots, and $h$ is the kernel bandwidth controlling the degree of smoothing. The Gaussian kernel function is selected in this work. The indicator function $\mathbb{I}_{\{x_{ij} \notin Masked\}}$ ensures that only the unmasked patches contribute to the estimation. The normalization constant $Z$ is defined as
\begin{equation}
Z = \sum_{i=1}^C \sum_{j=1}^N \mathbb{I}_{\{x_{ij} \notin Masked\}},
\end{equation}
where Z corresponds to the total number of unmasked patches. Hence, $\hat{f}_h(t)$ represents the estimated sampling density over time slots based solely on unmasked observations and is normalized to integrate to one.

\section{Experiments}
\label{sec4}

\subsection{Experiment setup}
\label{subsec4-1}

\textbf{Pretraining dataset and preprocessing.} CABrain was pretrained using the large-scale public dataset TUEG \cite{TUEG}. EEG recordings shorter than 5 min, along with the first and last minutes of each recording, were discarded to reduce low-quality segments. Nineteen common EEG channels from the 10--20 system were selected. The corresponding signals were band-pass filtered (0.3–75 Hz) and notch filtered (60 Hz), resampled to 200 Hz, and segmented into non-overlapping 30-second windows. Samples containing values exceeding ±100 µV were automatically removed, and all signals were normalized by 100 µV to ensure that values mainly ranged between –1 and 1.

\textbf{Pretraining.}
Because generating adaptive masks using nonparametric KDE is computationally expensive, 50,000 masks were generated and randomly sampled during pretraining. Using a single-core AMD Ryzen Threadripper PRO 3975WX CPU, generating 50,000 masks required approximately 5 h. If all 44,385,280 masks required for 40 epochs were to be generated during training, nearly 4,400 h would be consumed.  Therefore, pre-generation and reuse of masks are crucial. The two proposed methods, GCN-based CPE and adaptive masking, can be applied independently during pretraining; however, caution is required when applying them simultaneously. Adaptive masking can simultaneously mask various patches within specific time slots, increasing the number of cases in which the input nodes to the GCN have feature vectors consisting entirely of zeros. This may hinder the construction of informative positional embeddings. Therefore, we first pretrained a model with adaptive masking alone and subsequently applied GCN-based CPE through transfer learning to build the foundation model.

\textbf{Test datasets.} Experiments were conducted on 10 publicly available EEG datasets, as summarized in Table 1.

{\footnotesize
\begin{longtable}{cccc}

\caption{Overview of test datasets.}
\label{table1}\\

\toprule
Downstream task & Dataset & \# Channels & \# Class \\
\midrule
\endfirsthead

\caption[]{Overview of test datasets. (continued)} \\
\toprule
Downstream task & Dataset & \# Channels & \# Class  \\
\midrule
\endhead

\bottomrule
\endfoot

\bottomrule

\endlastfoot

\multirow{1}{*}{Mental stress detection} 
& MentalArithmetic \cite{MentalArithmetic}  & 20 & 2 \\

\multirow{1}{*}{Mental disorder diagnosis} 
& Mumtaz2016 \cite{Mumtaz} & 19 & 2 \\

\multirow{3}{*}{Motor Imagery classification} 
& SHU-MI \cite{SHU-MI} & 32 & 2  \\
& PhysioNet-MI \cite{PhySioNet} & 64  & 4  \\
& BCIC-IV-2a \cite{BCIC-IV-2a}  & 22 & 4  \\

\multirow{1}{*}{Seizure detection} 
& CHB-MIT \cite{CHB-MIT} & 16 & 2 \\

\multirow{2}{*}{Emotion recognition} 
& FACED \cite{FACED}  & 32 &  9 \\
& SEED-V \cite{SEED-V} & 62 &  5  \\

\multirow{1}{*}{Sleep stage classification} 
& ISRUC \cite{ISRUC} & 6  & 5 \\

\multirow{1}{*}{Imagined speech classification} 
& BCIC2020-3 \cite{BCIC2020} & 64 & 5  \\

\bottomrule

\end{longtable}
}

\textbf{Baseline methods.} The performance of CABrain was compared with both non-foundation and foundation models. Among the non-foundation models, representative CNN- and transformer-based architectures were used, i.e., EEGNet \cite{EEGNet} and CNN-Transformer \cite{CNN-Transformer}, respectively. Recent advances in EEG foundation models have demonstrated the potential of large-scale pretraining for learning generic EEG representations. Among EEG foundation models, BIOT \cite{BIOT}, LaBraM \cite{LaBraM}, CBraMod \cite{CBraMod}, and EEGMamba \cite{EEGMamba} were considered as baselines. CBraMod \cite{CBraMod} and EEGMamba \cite{EEGMamba} are among the most recently proposed methods that have achieved state-of-the-art performance on the test datasets used in this study.

\textbf{Metrics.} To ensure a fair comparison with previous studies, the following evaluation metrics were applied: For binary classification, the performance was assessed using balanced accuracy, area under the precision--recall curve (AUC-PR), and area under the receiver operating characteristic (AUROC), with AUROC designated as the monitoring score. For multi-class classification, balanced accuracy, Cohen’s kappa, and weighted F1 were employed, with Cohen’s kappa serving as the monitoring score. Performance metrics are reported as the mean ± standard deviation over five runs with different random seeds.

\textbf{Model architecture and implementation.} The proposed GCN-based CPE consisted of two GCN layers and was implemented using PyTorch Geometric. The input and output dimensions of the GCNConv layers were identical to the patch embedding dimension, and a rectified linear unit (ReLU) activation function was applied between the two GCNConv layers. The nonparametric KDE for adaptive mask generation was implemented using the gaussian\_kde function provided by the SciPy library. The structures of the classification heads were as follows: For the ISRUC and BCIC-IV-2a datasets, a single fully connected layer was employed; for the BCIC2020-3 and CHB-MIT datasets, two fully connected layers were set; and three fully connected layers were adopted for the remaining datasets. Binary cross-entropy loss was adopted for binary classification, whereas cross-entropy loss was used for multi-class classification.

\subsection{Experimental results}
\label{subsec4-2}

The hyperparameters used were identical to those for CBraMod \cite{CBraMod}, except for the learning rate. For each test dataset, the learning rate was selected using the validation set under the first random seed, and the best-performing values were used for experiments with the remaining seeds.

{\footnotesize
\begin{longtable}{llll}

\caption{Learning rates for test datasets.}
\label{table2}\\

\toprule
Dataset & CABrain(PE) & CABrain(Mask) & CABrain(Comb) \\
\midrule
\endfirsthead

\caption[]{Learning rates for test datasets. (continued)} \\
\toprule
Dataset & CABrain(PE) & CABrain(Mask) & CABrain(Comb) \\
\midrule
\endhead

\bottomrule
\endfoot

\bottomrule

\endlastfoot

 MentalArithmetic & 0.00001 & 0.000025 & 0.000025  \\
 Mumtaz2016 & 0.00005 & 0.000025 & 0.000025 \\
 SHU-MI & 0.0001 & 0.0001  & 0.000005  \\
 PhysioNet-MI & 0.00025 & 0.00025 & 0.00025 \\
 BCIC-IV-2a & 0.00025 & 0.00025  & 0.0025 \\
 CHB-MIT & 0.00005 & 0.000075 & 0.0001  \\
 FACED & 0.00025 & 0.00025 & 0.0005  \\
 SEED-V & 0.00025 & 0.00025 & 0.00001 \\
 ISRUC & 0.000025 & 0.000025 & 0.0001  \\
 BCIC2020-3 & 0.00025 & 0.00025 & 0.00025  \\

\bottomrule

\end{longtable}
}

The experimental results presented in Tables 3 and 4 were obtained by fine-tuning both the pretrained parameters of the foundation model and the parameters of the classifier on the test datasets. Values for the best-performing model are boldfaced, whereas those for the second-best model are underlined. Because EEGMamba \cite{EEGMamba} reported results for only five of the ten datasets, its performance was compared only for those datasets. CABrain(PE) replaces positional encoding in the baseline model CBraMod \cite{CBraMod} with the proposed GCN-based CPE. CABrain(Mask) applies adaptive mask generation during the pretraining stage but maintains the same network architecture as CBraMod. CABrain(Comb) integrates both methods.

The performance of GCN-based CPE is influenced by the number of edges connected to each channel-specific graph node, which is determined by the strength of the inter-channel correlations. We constructed a graph by connecting each node to its three most strongly correlated nodes. The experimental results reported below are based on this graph structure. Proper selection of the kernel bandwidth plays a critical role in nonparametric KDE. Thus, in the subsequent experiments, we set the kernel bandwidth to 0.5. In addition, the masking ratio of the input patches influences the performance. For these experiments, we employed a model pretrained with a 50\% masking ratio. The basis for selecting these parameters is discussed in Section 4.3.

{\footnotesize
\begin{longtable}{l l ccc}

\caption{Performance comparison with existing methods (binary-class).}
\label{table3}\\

\toprule
Dataset & Method & Bal. Acc. & AUC-PR & AUROC \\
\midrule
\endfirsthead

\caption[]{Performance comparison with existing methods (2-class). (continued)} \\
\toprule
Dataset & Method & Bal. Acc. & AUC-PR & AUROC \\
\midrule
\endhead

\bottomrule
\endfoot

\bottomrule

\endlastfoot

\multirow[t]{10}{*}{MentalArithmetic} 
& EEGNet            & $67.70\pm1.16$ & $57.63\pm1.02$ & $73.21\pm1.08$ \\
& CNN-Trans.   & $67.79\pm2.68$ & $57.77\pm2.85$ & $72.58\pm3.36$ \\
\cmidrule(l){2-5}
& BIOT              & $68.75\pm1.86$ & $60.04\pm1.95$ & $75.36\pm1.44$ \\
& LaBraM-Base       & $69.09\pm1.25$ & $59.99\pm1.55$ & $77.21\pm0.93$ \\
& CBraMod           & $72.56\pm1.32$ & $62.67\pm0.99$ & $79.05\pm0.73$ \\
\cmidrule(l){2-5}
& CABrain(PE)       & $73.47\pm2.52$ & $\underline{74.96}\pm2.39$ & $83.50\pm3.65$ \\
& CABrain(Mask)     & $\underline{77.71}\pm1.83$ & $70.23\pm3.54$ & $\underline{84.85}\pm1.18$ \\
& CABrain(Comb)           & $\textbf{79.58}\pm1.42$ & $\textbf{78.05}\pm2.83$ & $\textbf{85.00}\pm2.24$ \\
\midrule

\multirow[t]{10}{*}{Mumtaz2016} 
& EEGNet            & $92.32\pm1.04$ & $96.26\pm0.95$ & $96.39\pm0.93$ \\
& CNN-Trans.   & $93.05\pm0.68$ & $97.57\pm0.74$ & $97.42\pm0.59$ \\
\cmidrule(l){2-5}
& BIOT              & $93.58\pm0.52$ & $97.36\pm0.34$ & $97.58\pm0.42$ \\
& LaBraM-Base       & $94.09\pm0.79$ & $97.98\pm0.93$ & $97.82\pm0.57$ \\
& CBraMod           & $95.60\pm0.56$ & $\underline{99.23}\pm0.32$ & $\underline{99.21}\pm0.25$ \\
\cmidrule(l){2-5}
& CABrain(PE)       & $88.72\pm0.49$ & $97.09\pm0.19$ & $96.88\pm0.27$ \\
& CABrain(Mask)     & $\textbf{96.55}\pm0.32$ & $\textbf{99.56}\pm0.08$ & $\textbf{99.55}\pm0.08$ \\
& CABrain(Comb)           & $\underline{95.82}\pm0.63$ & $96.89\pm1.50$ & $98.79\pm0.49$ \\
\midrule

\multirow[t]{10}{*}{SHU-MI} 
& EEGNet            & $58.89\pm1.77$ & $63.11\pm1.42$ & $62.83\pm1.52$ \\
& CNN-Trans.   & $59.75\pm1.69$ & $64.12\pm0.76$ & $63.23\pm0.82$ \\
\cmidrule(l){2-5}
& BIOT              & $61.79\pm1.83$ & $67.70\pm1.19$ & $66.09\pm1.27$ \\
& LaBraM-Base       & $61.66\pm1.92$ & $67.61\pm0.83$ & $66.04\pm0.91$ \\
& CBraMod           & $\underline{63.70}\pm1.51$ & $\textbf{71.39}\pm0.88$ & $\textbf{69.88}\pm0.68$ \\
\cmidrule(l){2-5}
& CABrain(PE)       & $62.82\pm0.73$ & $68.81\pm1.30$ & $68.77\pm1.14$ \\
& CABrain(Mask)     & $\textbf{63.93}\pm0.68$ & $69.39\pm0.53$ & $\underline{69.50}\pm0.83$ \\
& CABrain(Comb)           & $63.62\pm0.81$ & $\underline{68.99}\pm0.92$ & $69.44\pm0.64$ \\
\midrule

\multirow[t]{10}{*}{CHB-MIT} 
& EEGNet            & $56.58\pm1.06$ & $19.14\pm1.82$ & $80.48\pm1.36$ \\
& CNN-Trans.   & $63.89\pm0.67$ & $24.79\pm2.27$ & $86.62\pm0.82$ \\
\cmidrule(l){2-5}
& BIOT              & $70.68\pm4.57$ & $32.77\pm4.60$ & $87.61\pm2.84$ \\
& LaBraM-Base       & $70.75\pm3.58$ & $32.87\pm4.02$ & $86.79\pm1.99$ \\
& CBraMod           & $73.98\pm2.84$ & $36.89\pm3.82$ & $88.92\pm1.54$ \\
& EEGMamba          & $74.45\pm2.46$ & $38.85\pm4.18$ & $89.38\pm1.61$ \\
\cmidrule(l){2-5}
& CABrain(PE)       & $\underline{74.62}\pm4.32$ & $55.50\pm12.46$ & $92.11\pm2.88$ \\
& CABrain(Mask)     & $71.94\pm3.68$ & $\textbf{58.01}\pm10.51$ & $\textbf{93.64}\pm2.70$ \\
& CABrain(Comb)           & $\textbf{76.87}\pm4.41$ & $\underline{57.54}\pm11.34$ & $\underline{92.72}\pm1.02$ \\
\bottomrule

\end{longtable}
}

{\footnotesize
\begin{longtable}{l l ccc}

\caption{Performance comparison with existing methods (multi-class).}
\label{table4}\\

\toprule
Dataset & Method & Bal. Acc. & Cohen’s Kappa & Weighted F1 \\
\midrule
\endfirsthead

\caption[]{Performance comparison with existing methods (multi-class). (continued)} \\
\toprule
Dataset & Method & Bal. Acc. & Cohen’s Kappa & Weighted F1 \\
\midrule
\endhead

\bottomrule
\endfoot

\bottomrule

\endlastfoot

\multirow[t]{10}{*}{PhysioNet-MI} 
& EEGNet            & $58.14\pm1.25$ & $44.68\pm1.99$ & $57.96\pm1.15$ \\
& CNN-Trans.   & $60.53\pm1.18$ & $47.25\pm2.23$ & $60.41\pm1.05$ \\
\cmidrule(l){2-5}
& BIOT              & $61.53\pm1.54$ & $48.75\pm2.72$ & $61.58\pm1.97$ \\
& LaBraM-Base       & $61.73\pm1.22$ & $49.12\pm1.92$ & $61.77\pm1.41$ \\
& CBraMod           & $\underline{64.17}\pm0.91$ & $\underline{52.22}\pm1.69$ & $\underline{64.27}\pm1.00$ \\
& EEGMamba          & $63.23\pm0.84$ & $51.31\pm1.24$ & $63.42\pm0.98$ \\
\cmidrule(l){2-5}
& CABrain(PE)       & $63.26\pm0.61$ & $51.02\pm0.81$ & $63.30\pm0.59$ \\
& CABrain(Mask)     & $\textbf{64.33}\pm0.40$ & $\textbf{52.44}\pm0.53$ & $\textbf{64.37}\pm0.37$ \\
& CABrain(Comb)           & $64.02\pm0.54$ & $52.02\pm0.72$ & $64.05\pm0.54$ \\
\midrule

\multirow[t]{10}{*}{BCIC-IV-2a} 
& EEGNet            & $44.82\pm0.94$ & $26.93\pm1.21$ & $42.26\pm1.08$ \\
& CNN-Trans.   & $46.00\pm1.08$ & $28.00\pm1.48$ & $44.60\pm1.14$ \\
\cmidrule(l){2-5}
& BIOT              & $47.48\pm0.93$ & $29.97\pm1.39$ & $46.07\pm1.25$ \\
& LaBraM-Base       & $48.69\pm0.85$ & $31.59\pm1.54$ & $47.58\pm1.03$ \\
& CBraMod           & $51.38\pm0.66$ & $35.18\pm0.94$ & $49.84\pm0.85$ \\
\cmidrule(l){2-5}
& CABrain(PE)       & $\underline{55.40}\pm0.62$ & $\underline{40.53}\pm0.82$ & $\underline{55.00}\pm0.76$ \\
& CABrain(Mask)     & $54.24\pm2.62$ & $38.98\pm3.49$ & $53.32\pm3.28$ \\
& CABrain(Comb)           & $\textbf{59.15}\pm1.27$ & $\textbf{45.53}\pm1.70$ & $\textbf{58.83}\pm1.13$ \\
\midrule

\multirow[t]{10}{*}{FACED} 
& EEGNet            & $40.90\pm1.22$ & $33.42\pm2.51$ & $41.24\pm1.41$ \\
& CNN-Trans.   & $46.97\pm1.32$ & $40.17\pm1.68$ & $47.20\pm1.25$ \\
\cmidrule(l){2-5}
& BIOT              & $51.18\pm1.18$ & $44.76\pm2.54$ & $51.36\pm1.12$ \\
& LaBraM-Base       & $52.73\pm1.07$ & $46.98\pm1.88$ & $52.88\pm1.02$ \\
& CBraMod           & $55.09\pm0.89$ & $50.41\pm1.22$ & $56.18\pm0.93$ \\
& EEGMamba          & $56.46\pm0.96$ & $51.34\pm1.27$ & $56.73\pm0.76$ \\
\cmidrule(l){2-5}
& CABrain(PE)       & $\underline{58.08}\pm0.50$ & $\underline{52.49}\pm0.56$ & $\underline{58.32}\pm0.53$ \\
& CABrain(Mask)     & $58.08\pm0.80$ & $52.47\pm0.92$ & $58.26\pm0.77$ \\
& CABrain(Comb)           & $\textbf{58.63}\pm0.91$ & $\textbf{53.21}\pm0.99$ & $\textbf{59.27}\pm0.75$ \\
\midrule

\multirow[t]{10}{*}{SEED-V} 
& EEGNet            & $29.61\pm1.02$ & $10.06\pm1.43$ & $27.49\pm0.98$ \\
& CNN-Trans.   & $36.78\pm0.78$ & $20.72\pm1.83$ & $36.42\pm0.88$ \\
\cmidrule(l){2-5}
& BIOT              & $38.37\pm1.87$ & $22.61\pm2.62$ & $38.56\pm2.03$ \\
& LaBraM-Base       & $39.76\pm1.38$ & $23.86\pm2.09$ & $39.74\pm1.11$ \\
& CBraMod           & $\underline{40.91}\pm0.97$ & $25.69\pm1.43$ & $41.01\pm1.08$ \\
\cmidrule(l){2-5}
& CABrain(PE)       & $40.57\pm0.07$ & $\underline{26.47}\pm0.16$ & $\underline{41.60}\pm0.12$ \\
& CABrain(Mask)     & $40.45\pm0.13$ & $26.25\pm0.14$ & $41.40\pm0.10$ \\
& CABrain(Comb)           & $\textbf{40.95}\pm0.21$ & $\textbf{26.80}\pm0.34$ & $\textbf{41.85}\pm0.31$ \\
\midrule

\multirow[t]{10}{*}{ISRUC} 
& EEGNet            & $71.54\pm1.21$ & $70.40\pm1.73$ & $75.13\pm1.24$ \\
& CNN-Trans.   & $73.63\pm0.87$ & $71.29\pm1.21$ & $77.19\pm1.05$ \\
\cmidrule(l){2-5}
& BIOT              & $75.27\pm1.21$ & $71.92\pm2.31$ & $77.90\pm1.46$ \\
& LaBraM-Base       & $76.33\pm1.02$ & $72.31\pm1.82$ & $78.10\pm1.33$ \\
& CBraMod           & $78.65\pm1.10$ & $74.42\pm1.52$ & $80.11\pm0.99$ \\
& EEGMamba          & $\underline{78.99}\pm0.93$ & $74.86\pm1.22$ & $80.54\pm0.76$ \\
\cmidrule(l){2-5}
& CABrain(PE)       & $\textbf{79.19}\pm0.32$ & $\textbf{75.71}\pm0.67$ & $\textbf{81.41}\pm0.49$ \\
& CABrain(Mask)     & $78.73\pm0.23$ & $74.61\pm0.24$ & $80.31\pm0.16$ \\
& CABrain(Comb)           & $78.71\pm0.40$ & $\underline{75.36}\pm0.63$ & $\underline{80.79}\pm0.55$ \\
\midrule

\multirow{10}{*}{BCIC2020-3} 
& EEGNet            & $44.13\pm0.96$ & $30.16\pm1.23$ & $44.13\pm1.02$ \\
& CNN-Trans.   & $45.33\pm0.92$ & $31.66\pm1.18$ & $45.06\pm1.27$ \\
\cmidrule(l){2-5}
& BIOT              & $49.20\pm0.86$ & $36.50\pm1.76$ & $49.17\pm0.79$ \\
& LaBraM-Base       & $50.60\pm1.55$ & $38.00\pm2.42$ & $50.54\pm2.05$ \\
& CBraMod           & $53.73\pm1.08$ & $42.16\pm1.63$ & $53.83\pm0.96$ \\
& EEGMamba          & $53.05\pm1.09$ & $41.83\pm1.57$ & $53.47\pm0.82$ \\
\cmidrule(l){2-5}
& CABrain(PE)       & $\underline{55.79}\pm1.76$ & $\underline{44.73}\pm2.20$ & $\underline{55.80}\pm1.76$ \\
& CABrain(Mask)     & $54.35\pm2.27$ & $42.93\pm2.83$ & $54.31\pm2.31$ \\
& CABrain(Comb)           & $\textbf{57.04}\pm0.62$ & $\textbf{46.30}\pm0.78$ & $\textbf{57.05}\pm0.62$ \\
\bottomrule

\end{longtable}
}

\begin{figure}[H]
\centering
\includegraphics[width=1\textwidth]{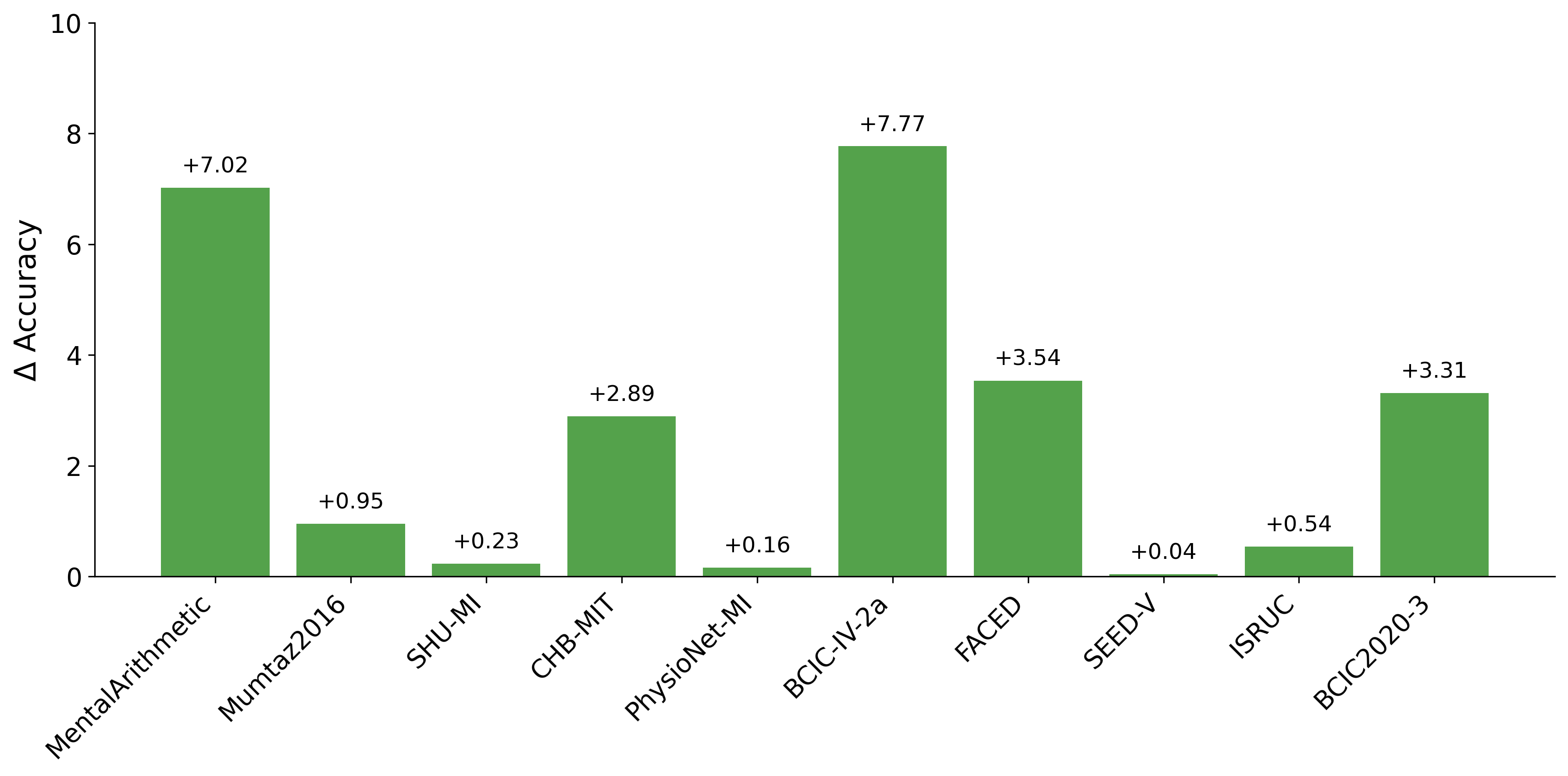}
\caption{Relative accuracy of the best-performing CABrain variant compared to CBraMod.}\label{fig7}
\end{figure}

When comparing the performance of different methods across the three metrics for each test dataset, similar trends were observed. In the interest of conciseness, the following discussion focuses on balanced accuracy. Among the baseline methods, CBraMod \cite{CBraMod} achieved the best performance on seven datasets, whereas EEGMamba \cite{EEGMamba} achieved the best performance on three datasets. As CBraMod achieved the best performance on the largest number of datasets among the baseline methods, we examined the extent to which the proposed methods surpassed it.

CABrain(PE), CABrain(Mask), and CABrain(Comb) demonstrated superior performances over CBraMod on six, eight, and eight test datasets, respectively. Figure 7 illustrates the performance gap between the best-performing CABrain variant and CBraMod for each test dataset, indicating that at least one CABrain variant outperformed CBraMod across all datasets. Because of the diverse characteristics of the datasets, none of the CABrain variants consistently achieved the best performance. However, CABrain(Comb) generally achieved the best or second-best results, highlighting the effectiveness of combining the two proposed approaches. CABrain(Comb) achieved an average improvement of 2.46 percentage points in classification accuracy across the 10 test datasets. Notably, the most recent baseline, EEGMamba, outperformed CBraMod by only 0.11 percentage points on average, highlighting the substantial performance gains achieved by CABrain(Comb).

{\footnotesize
\begin{longtable}{l cccc}

\caption{Average balanced accuracy across all, binary, and multi-class datasets.}
\label{table5}\\

\toprule
Dataset Group & CABrain(PE) & CABrain(Mask) & CABrain(Comb) & CBraMod \\
\midrule
\endfirsthead

\caption[]{Average balanced accuracy across all, binary, and multi-class datasets. (continued)} \\
\toprule
Dataset Group & CABrain(PE) & CABrain(Mask) & CABrain(Comb) & CBraMod \\
\midrule
\endhead

\bottomrule
\endfoot

\bottomrule
\endlastfoot

All datasets     & 65.19 & \underline{66.03} & \textbf{67.44} & 64.98 \\
Binary-class     & 74.91 & \underline{77.53} & \textbf{78.97} & 76.46 \\
Multi-class      & \underline{58.72} & 58.36 & \textbf{59.75} & 57.32 \\

\bottomrule

\end{longtable}
}

To better understand the relative contributions of the two proposed approaches, we further analyzed their performance trends across binary- and multi-class downstream tasks. Although the experimental results did not indicate that either GCN-based positional encoding or adaptive masking is universally superior, with their performance depending on the number of classes, clear trends could be observed.

For the binary-class tasks, CABrain(PE) outperformed CABrain(Mask) only on CHB-MIT, whereas CABrain(Mask) achieved higher balanced accuracy on the remaining three datasets. In contrast, for the multi-class tasks, CABrain(PE) achieved superior or more stable performance on five of the six datasets, whereas CABrain(Mask) outperformed CABrain(PE) only on PhysioNet-MI. On FACED, although both methods achieved the same balanced accuracy, CABrain(PE) exhibited a smaller standard deviation, indicating more stable performance.

These trends were also reflected in the average balanced accuracy (Table 5). For the binary-class tasks, CABrain(Mask) achieved a higher average balanced accuracy than CABrain(PE) (77.53\% vs. 74.91\%). In contrast, for the multi-class tasks, CABrain(PE) achieved an average balanced accuracy of 58.72\%, slightly outperforming CABrain(Mask) (58.36\%). Furthermore, CABrain(PE) outperformed CBraMod by 1.4 percentage points on the multi-class datasets.

\begin{figure}[H]
\centering
\includegraphics[width=1\textwidth]{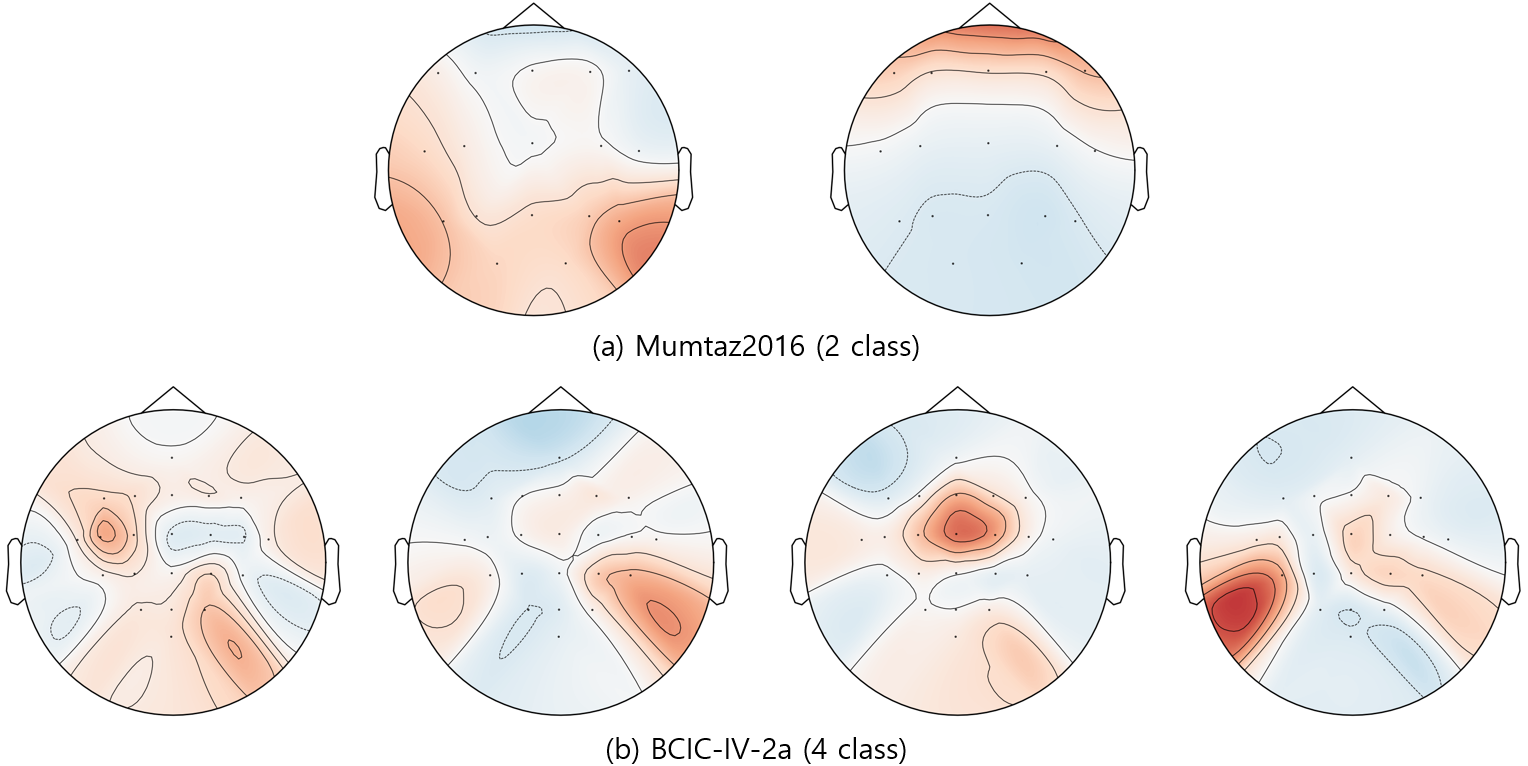}
\caption{Class-wise EEG topographies (Mumtaz2016 and BCIC-IV-2a).}\label{fig8}
\end{figure}

These observations suggest a possible relationship between the number of classes and relative utility of the two proposed components, although other dataset characteristics may also contribute to the observed differences. In many binary-class EEG tasks, such as mental stress detection and mental disorder diagnosis, the primary objective is to distinguish between two brain states. Compared with multi-class EEG decoding, the distinction between classes is simpler. This aspect is reflected in Figure 8(a), where the two classes in Mumtaz2016 exhibit distinct global topographic patterns. Such coarse-grained differences can often be identified from broader temporal and inter-channel contextual information without precise modeling of localized channel-wise spatial relationships. The proposed adaptive masking strategy encourages the model to leverage weakly correlated channels and temporal dependencies during masked patch reconstruction, thereby preventing excessive reliance on a subset of channels.

In contrast, multi-class EEG decoding generally involves more complicated decision boundaries, requiring finer discrimination among multiple classes. As illustrated in Figure 8(b), the four classes in BCIC-IV-2a exhibit more subtle and localized differences in their EEG topographies. Distinguishing these classes requires preserving fine-grained spatial relationships among channels, making accurate positional information more beneficial for multi-class EEG decoding. Under these settings, preserving subtle class-specific spatial relationships among EEG channels may become critical. The proposed GCN-based CPE approach explicitly preserves correlation-aware spatial topology, enabling the model to effectively exploit localized discriminative channel relationships.

Overall, the results suggest that the relative utility of the two components may depend on the nature of the classification task. Adaptive masking appears to be particularly beneficial when coarse-grained differences between classes can be captured from broader contextual information, whereas GCN-based CPE may provide greater benefits when accurate modeling of subtle spatial relationships among EEG channels is required. The observed tendency may help explain why adaptive masking achieved stronger performance on several binary-class datasets, whereas GCN-based positional encoding performed more favorably on many multi-class datasets.

\subsection{Ablation study}
\label{subsec4-3}

Ablation studies were conducted to evaluate the effects of key hyperparameters in the proposed framework. Specifically, we investigated four key hyperparameters that affect the performance of the foundation model: the degree of a node (node degree) employed to construct the graph structure for GCN-based CPE, masking ratio and kernel bandwidth of the KDE-based adaptive masking strategy, and size of the mask library used for masking pattern sampling.\\

\textbf{Node Degree.}
The degree of a node in a graph refers to the number of edges connected to that node. To examine the impact of the node degree on GCN-based CPE, the node degree was varied from one to five. Table 6 reports the balanced accuracy of the test datasets for the five node degrees. Although the optimal degree varied across datasets, a certain trend was observed. To examine the overall effect of node degree, the average relative accuracy was examined. For each dataset, the mean accuracy was first computed by averaging the accuracies obtained with the five node degrees. The relative accuracy for a given node degree was then calculated as the difference between the accuracy achieved at that degree and the corresponding dataset-specific mean accuracy. Lastly, the average relative accuracy for each node degree was obtained by averaging the relative accuracies across the 10 datasets.

{\footnotesize
\begin{longtable}{l ccccc}

\caption{Balanced accuracy across different node degrees.}
\label{table6}\\

\toprule
Dataset & \multicolumn{5}{c}{Node Degree} \\
\cmidrule(lr){2-6}
 & 1 & 2 & 3 & 4 & 5\\
\midrule
\endfirsthead

\caption[]{Balanced accuracy across different node degrees. (continued)} \\
\toprule
Dataset & \multicolumn{5}{c}{Node Degree} \\
\cmidrule(lr){2-6}
 & 1 & 2 & 3 & 4 & 5 \\
\midrule
\endhead

\bottomrule
\endfoot

\bottomrule

\endlastfoot

MentalArithmetic & 64.65 & \textbf{75.00} & 73.47 & 69.72 & 71.25 \\
Mumtaz2016       & 86.07 & 87.40 & \textbf{88.72} & 87.90 & 87.55 \\
SHU-MI           & 61.31 & 61.29 & \textbf{62.82} & 61.84 & 60.30 \\
PhysioNet-MI     & 62.83 & 62.70 & \textbf{63.26} & 62.16 & 62.05 \\
BCIC-IV-2a       & \textbf{57.93} & 54.77 & 55.40 & 54.60 & 54.58 \\
CHB-MIT          & 63.00 & 70.23 & \textbf{74.62} & 72.18 & 69.90 \\
FACED            & 56.92 & 56.59 & \textbf{58.08} & 57.67 & 56.41 \\
SEED-V           & 38.93 & 39.05 & \textbf{40.57} & 39.93 & 39.58 \\
ISRUC            & 75.22 & 79.12 & \textbf{79.19} & 78.36 & 76.76 \\
BCIC2020-3       & 37.57 & 52.61 & 55.79 & \textbf{56.29} & 52.05 \\

\bottomrule

\end{longtable}
}

\begin{figure}[H]
\centering
\includegraphics[width=0.65\textwidth]{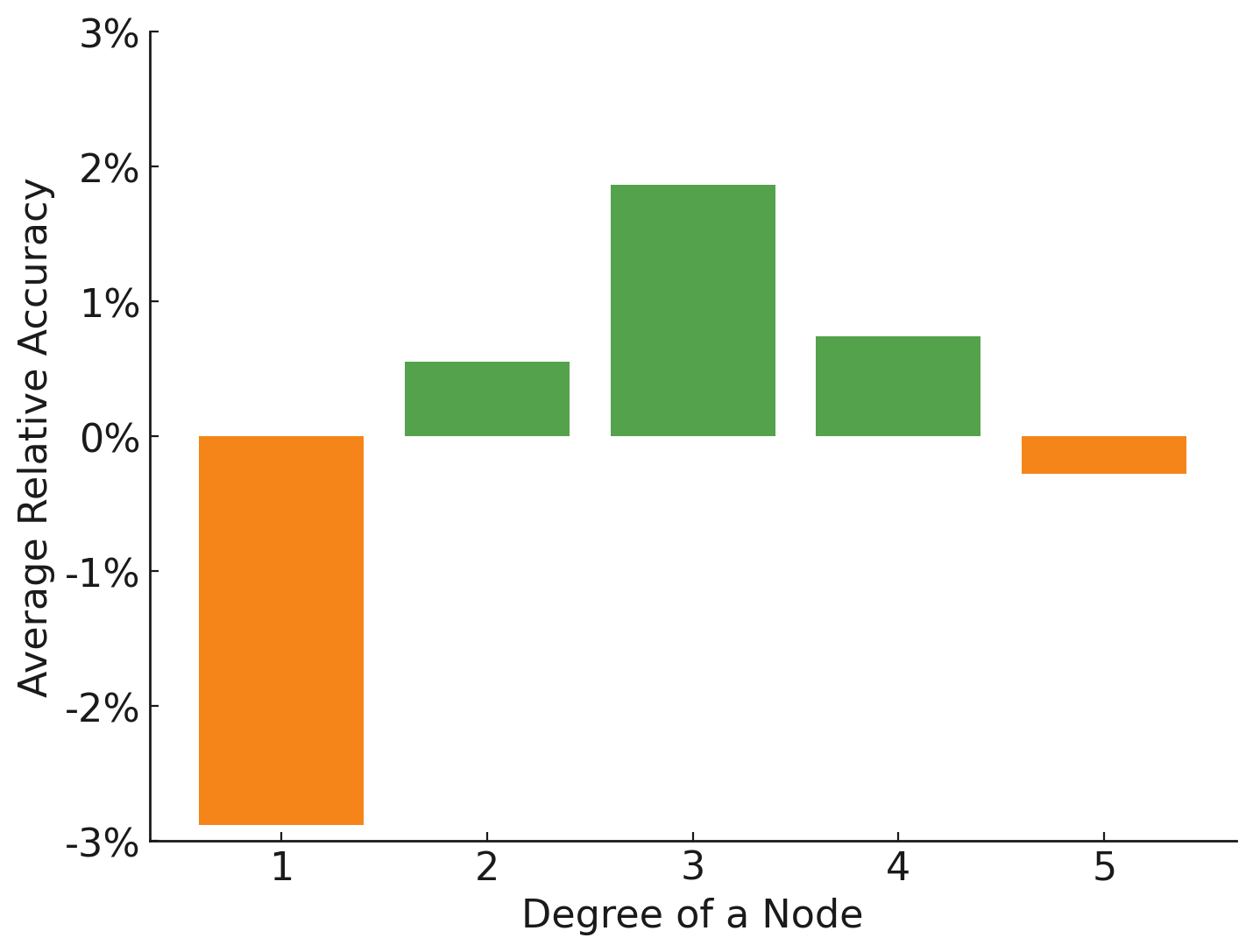}
\caption{Effect of node degree on classification accuracy.}\label{fig9} 
\end{figure}

The average relative accuracies for the five node degrees are shown in Figure 9. Moderate connectivity (three edges per node) resulted in the best performance. When each node was connected to a single strongly correlated node, the performance decreased by approximately 3\% relative to the average accuracy across datasets. This degradation is likely attributable to the fact that with only a single-edge connection, the model had no alternative nodes to compensate with when the signal of the connected node was corrupted by noise. Furthermore, if the connected node corresponded to a masked patch, the graph effectively lost that edge, rendering the GCN computation equivalent to operating on a disconnected structure. These factors explain the sharp decline in performance when the node degree was limited to one. Additionally, when the node degree reached four or higher, the performance gradually declined. This degradation can be attributed to excessive connectivity, which introduces redundancy and may dilute the discriminative power of graph representation.\\

\textbf{Masking ratio.}
To evaluate the impact of the masking ratio, experiments were conducted with masking ratios ranging from 30\% to 70\% in increments of 10\%. Table 7 presents the balanced accuracy for the datasets under five masking ratios. Although the masking ratio that yielded the best accuracy was not identical across all datasets, masking 50\% of the patches provided the highest accuracy for more datasets compared with any other masking ratio. This trend is evident in Figure 10, which compares the average relative accuracies. The effect of the random masking ratio on the performance has been reported in LaBraM \cite{LaBraM}, CBraMod \cite{CBraMod}, and EEGMamba \cite{EEGMamba}. Consistently, both adaptive and random masking achieved their best performances when the masking ratio was approximately 50\%.

The experimental results indicate that both excessive masking (70\%) and insufficient masking (30\%) diminish the effectiveness of representation learning. When the masking ratio is low, patch reconstruction becomes easier, because patches from strongly correlated channels are more likely to remain unmasked, limiting the sufficiency of representation learning. Conversely, when the masking ratio was considerably high, the model had limited observable context available for reconstruction. This shortage of information forces the model to make uncertain predictions and reduces the stability of training, which, in turn, lowers the classification accuracy.

{\footnotesize
\begin{longtable}{l ccccc}

\caption{Balanced accuracy across different masking ratios.}
\label{table7}\\

\toprule
Dataset & \multicolumn{5}{c}{Masking ratio} \\
\cmidrule(lr){2-6}
 & 30\% & 40\% & 50\% & 60\% & 70\%  \\
\midrule

\endfirsthead

\caption[]{Balanced accuracy across different masking ratios. (continued)} \\
\toprule
Dataset & \multicolumn{5}{c}{Masking ratio} \\
\cmidrule(lr){2-6}
 & 30\% & 40\% & 50\% & 60\% & 70\%  \\
\midrule
\endhead

\bottomrule
\endfoot

\bottomrule

\endlastfoot

MentalArithmetic & 69.44 & 70.63 & \textbf{77.71} & 70.21 & 68.54  \\
Mumtaz2016       & 81.56 & 90.37 & \textbf{96.55} & 88.82 & 84.81  \\
SHU-MI           & 61.14 & 61.83 & \textbf{63.93} & 61.65 & 60.50  \\
PhysioNet-MI     & 62.62 & 63.22 & \textbf{64.33} & 62.98 & 62.29  \\
BCIC-IV-2a       & 50.38 & 51.93 & \textbf{54.24} & 53.28 & 53.09 \\
CHB-MIT          & 64.34 & 65.60 & \textbf{71.94} & 67.45 & 62.73 \\
FACED            & 58.52 & \textbf{58.62} & 58.08 & 58.11 & 54.46 \\
SEED-V           & 39.67 & 39.33 & \textbf{40.45} & 40.05 & 39.60  \\
ISRUC            & 75.25 & 77.31 & \textbf{78.73} & 77.78 & 78.11  \\
BCIC2020-3       & 51.33 & 54.24 & \textbf{54.35} & 53.76 & 46.03  \\
\bottomrule

\end{longtable}
}

\begin{figure}[H]
\centering
\includegraphics[width=0.63\textwidth]{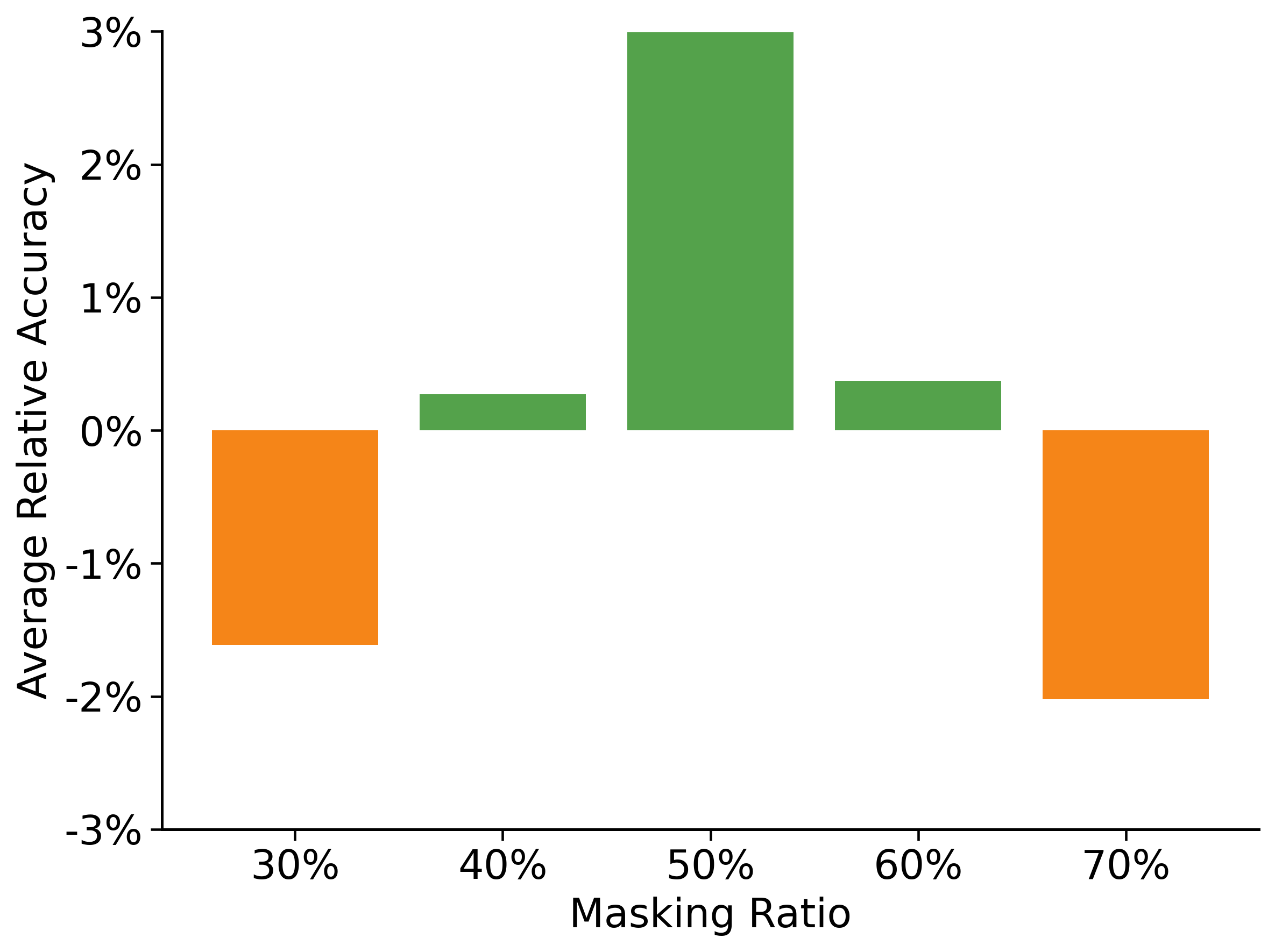}
\caption{Effect of masking ratio on classification accuracy.}\label{fig10}
\end{figure}

\textbf{Kernel bandwidth.}
In nonparametric KDE, the kernel bandwidth determines the range over which neighboring observations contribute to the estimated probability density. In adaptive masking, a small kernel bandwidth restricts the contribution of neighboring observations to the estimated probability density, resulting in sampling probabilities that more closely resemble those of random masking. As the kernel bandwidth increases, observations contribute to a broader range of neighboring time slots, increasing the likelihood that time slots containing previously masked patches are sampled. Therefore, a moderate kernel bandwidth (approximately 0.5) is a reasonable choice, providing a balance between under-smoothing at significantly small bandwidths and oversmoothing at larger bandwidths. As shown in Table 8, a kernel bandwidth of 0.5 does not guarantee the best performance across all datasets but yields the highest proportion of best-performing cases. This trend can be observed in Figure 11, which compares the average relative accuracies.

{\footnotesize
\begin{longtable}{l ccccc}

\caption{Balanced accuracy across different kernel bandwidths.}
\label{table8}\\

\toprule
Dataset & \multicolumn{5}{c}{Kernel bandwidth} \\
\cmidrule(lr){2-6}
 & 0.1 & 0.3 & 0.5 & 0.7 & 1  \\
\midrule
\endfirsthead

\caption[]{Balanced accuracy across different kernel bandwidths. (continued)} \\
\toprule
Dataset & \multicolumn{5}{c}{Kernel bandwidth} \\
\cmidrule(lr){2-6}
 & 0.1 & 0.3 & 0.5 & 0.7 & 1  \\
\midrule
\endhead

\bottomrule
\endfoot

\bottomrule

\endlastfoot

MentalArithmetic & 69.72 & 70.49 & \textbf{77.71} & 72.92 & 69.24 \\
Mumtaz2016       & 90.10 & 96.49 & \textbf{96.55} & 89.25 & 85.31  \\
SHU-MI           & 61.31 & 61.76 & \textbf{63.93} & 62.48 & 62.51  \\
PhysioNet-MI     & 62.75 & 62.63 & \textbf{64.33} & 63.21 & 62.52  \\
BCIC-IV-2a       & 52.62 & 53.32 & \textbf{54.24} & 54.18 & 50.61  \\
CHB-MIT          & 58.93 & 62.93 & \textbf{71.94} & 64.37 & 59.00  \\
FACED            & 57.64 & \textbf{58.65} & 58.08 & 58.11 & 58.30  \\
SEED-V           & 39.63 & 39.63 & \textbf{40.45} & 39.66 & 40.00  \\
ISRUC            & 77.24 & 77.41 & \textbf{78.73} & 78.06 & 77.50  \\
BCIC2020-3       & 53.17 & 53.81 & \textbf{54.35} & 53.41 & 49.47  \\
\bottomrule

\end{longtable}
}

\begin{figure}[H]
\centering
\includegraphics[width=0.63\textwidth]{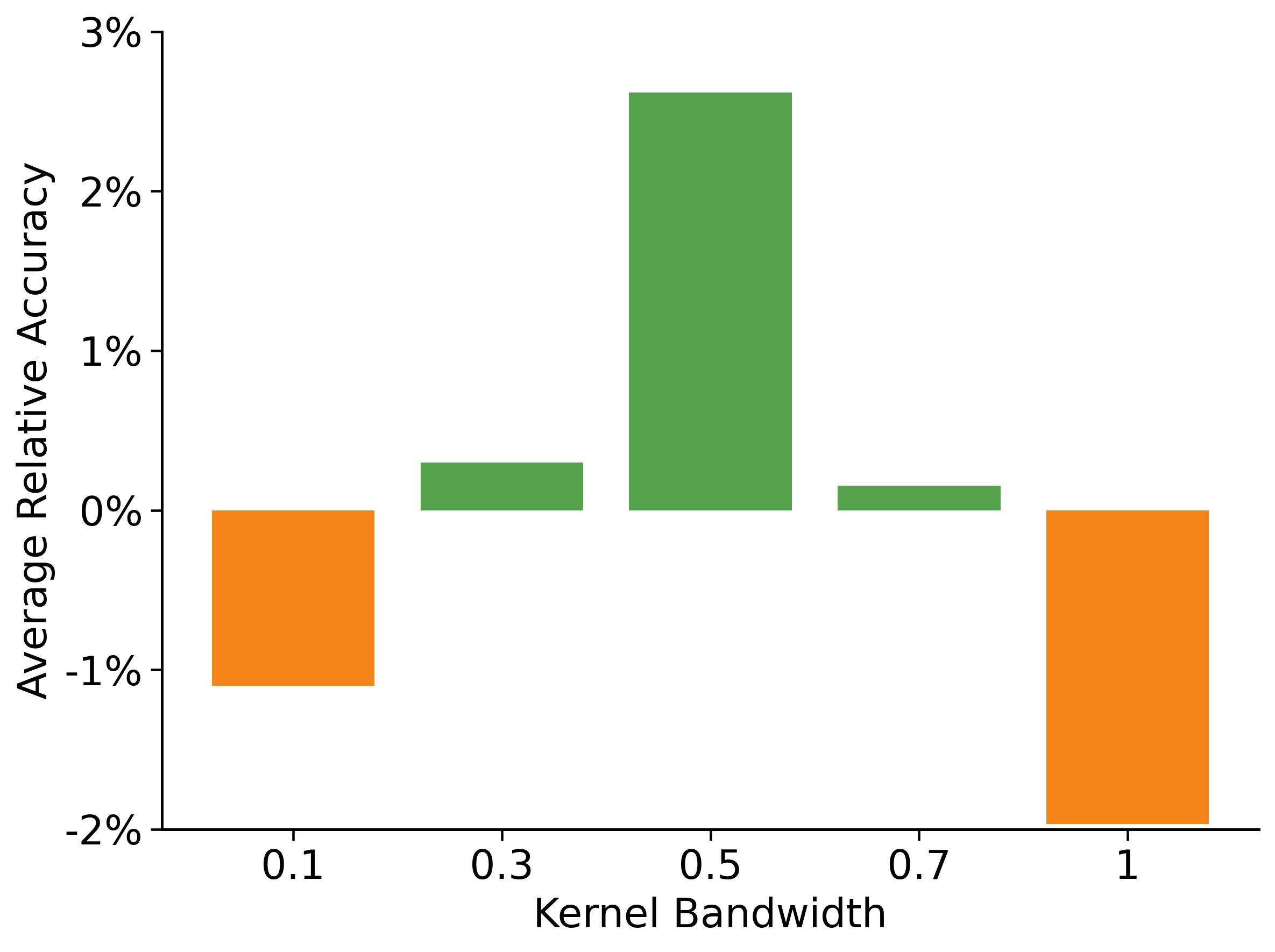}
\caption{Effect of kernel bandwidth on classification accuracy.}\label{fig11} 
\end{figure}

\textbf{Mask library size.}
The proposed adaptive masking strategy relies on a pre-generated mask library sampled from a KDE-based distribution. A natural question then is whether the library size is sufficiently large to represent the diversity implied by the estimated distribution. To investigate this issue, we examined libraries containing 5K, 10K, 50K, and 100K masks.

{\footnotesize
\begin{longtable}{l cccc}

\caption{Balanced accuracy under different mask library sizes.}
\label{table9}\\

\toprule
Dataset & \multicolumn{4}{c}{Mask library size} \\
\cmidrule(lr){2-5}
 & 5K & 10K & 50K & 100K \\
\midrule
\endfirsthead

\caption[]{Balanced accuracy under different mask library sizes. (continued)} \\
\toprule
Dataset & \multicolumn{4}{c}{Mask library size} \\
\cmidrule(lr){2-5}
 & 5K & 10K & 50K & 100K \\
\midrule
\endhead

\bottomrule
\endfoot

\bottomrule
\endlastfoot

MentalArithmetic & 64.23 & 75.63 & 77.71 & \textbf{78.13} \\
Mumtaz2016       & 82.38 & 95.78 & \textbf{96.55} & 95.55 \\
SHU-MI           & 59.45 & 62.82 & \textbf{63.93} & 63.41 \\
PhysioNet-MI     & 63.39 & 64.08 & \textbf{64.33} & 62.86 \\
BCIC-IV-2a       & 55.30 & 54.13 & 54.24 & \textbf{56.91} \\
CHB-MIT          & 71.73 & 71.45 & 71.94 & \textbf{72.18} \\
FACED            & 57.13 & 58.19 & 58.08 & \textbf{58.71} \\
SEED-V           & 39.59 & 40.13 & \textbf{40.45} & 40.29 \\
ISRUC            & 77.12 & 77.19 & \textbf{78.73} & 78.38 \\
BCIC2020-3       & 48.53 & 53.36 & \textbf{54.35} & 53.73 \\
\bottomrule

\end{longtable}
}

\begin{figure}[H]
\centering
\includegraphics[width=0.63\textwidth]{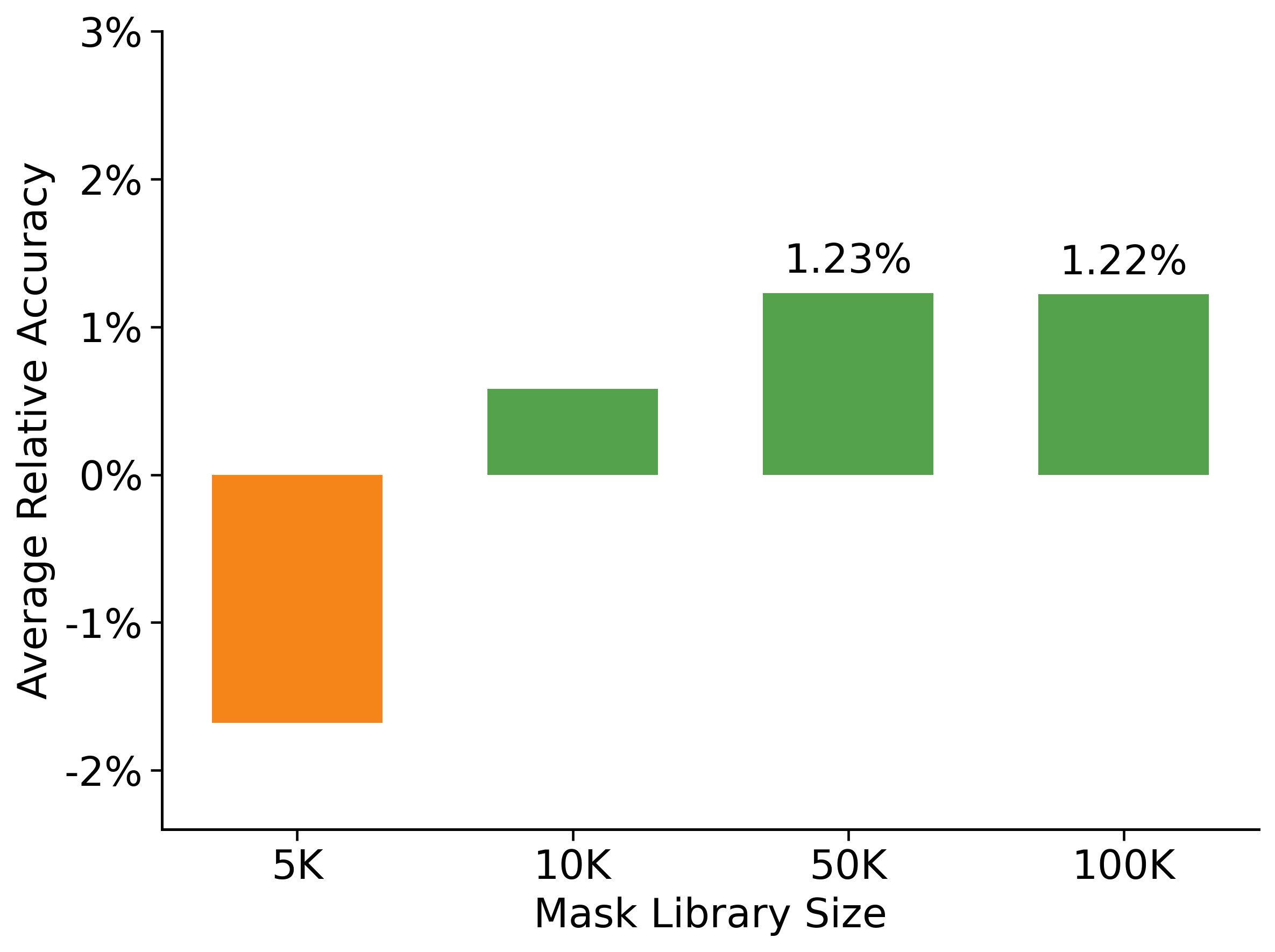}
\caption{Effect of mask library size on classification accuracy.}\label{fig12} 
\end{figure}

As shown in Table 9, performance improved as the library size increased to 50K, suggesting that additional mask diversity benefits representation learning. However, no performance improvement was observed when the library size was further increased to 100K. This trend can be observed in Figure 12, which compares the average relative accuracies. Given the additional computational cost required to generate larger libraries and the limited performance improvements beyond 50K masks, a size of 50K masks was adopted as the default setting.

\subsection{Validation of Design Choices}
\label{subsec4-4}

To justify the major design choices of CABrain, additional comparative experiments were conducted on both the masking strategy and GCN architecture. Specifically, we compare the proposed KDE-based adaptive masking strategy with several EEG-oriented masking baselines and evaluate alternative uses of GCN, including a GCN-only architecture and a feature-fusion approach that combines GCN outputs with transformer encoder features.. The objective of these experiments was to determine whether the proposed adaptive masking strategy and the use of GCN as CPE provide advantages over simpler alternatives.\\

\textbf{Comparison with EEG-Oriented Masking Strategies.}
To determine whether the proposed adaptive masking strategy provides benefits beyond those achievable by simple EEG-oriented masking strategies, we compare it with three masking baselines designed to exploit EEG characteristics. The first baseline, Channel-wise Masking, randomly selects EEG channels and masks all patches belonging to the selected channels. The second baseline, Time-slot Masking, randomly selects time slots and masks all patches within the selected time slots. The third baseline, Partial Time-slot Masking, randomly selects time slots and masks only a subset of patches within each selected time slot. For this baseline, the fraction of patches masked within each selected time slot is randomly sampled from predefined ranges (e.g., 10–90\% or 30–70\%).

Table 10 lists the balanced accuracies obtained using different masking strategies. CH, Time, PTS(30–70), PTS(10–90), RND, and Proposed denote Channel-wise Masking, Time-slot Masking, Partial Time-slot Masking with within-slot masking proportions of 30–70\% and 10–90\%, random masking and the proposed KDE-based adaptive masking strategy, respectively. All methods adopted the same masking ratio of 50\% to ensure a fair comparison.

{\footnotesize
\begin{longtable}{l cccccc}

\caption{Balanced accuracy under different masking strategies.}
\label{table10}\\

\toprule
Dataset & \multicolumn{6}{c}{Masking strategies} \\
\cmidrule(lr){2-7}
 & CH & Time & PTS(30–70) & PTS(10–90) & RND & Proposed \\
\midrule
\endfirsthead

\caption[]{Balanced accuracy under different masking strategies. (continued)} \\
\toprule
Dataset & \multicolumn{5}{c}{Mask library size} \\
\cmidrule(lr){2-7}
 & CH & Time & PTS(30–70) & PTS(10–90) & RND & Proposed \\
\midrule
\endhead

\bottomrule
\endfoot

\bottomrule
\endlastfoot

MentalArithmetic & 64.31 & 72.29 & 73.61 & 76.67 & 72.56 & \textbf{77.71} \\
Mumtaz2016       & 89.25 & 89.79 & 88.30 & 89.82 & 95.60 & \textbf{96.55} \\
SHU-MI           & 61.22 & 61.07 & 61.26 & 61.32 & 63.70 & \textbf{63.93} \\
PhysioNet-MI     & 62.75 & 65.05 & 61.79 & 63.67 & 64.17 & \textbf{64.33} \\
BCIC-IV-2a       & 53.73 & 55.50 & \textbf{56.16} & 55.47 & 51.38 & 54.24 \\
CHB-MIT          & 69.42 & 66.29 & 71.21 & 69.51 & \textbf{73.98} & 71.94 \\
FACED            & 55.38 & \textbf{60.07} & 54.32 & 54.43 & 55.09 & 58.08 \\
SEED-V           & 39.60 & 39.75 & 40.11 & 39.96 & \textbf{40.91} & 40.45 \\
ISRUC            & 72.55 & 71.73 & 76.00 & 76.51 & 78.65 & \textbf{78.73} \\
BCIC2020-3       & 48.19 & 52.37 & 47.04 & 51.84 & 53.73 & \textbf{54.35} \\

\midrule
Average          & 61.64 & 63.39 & 62.98 & 63.92 & 64.98 & \textbf{66.03} \\
\bottomrule

\end{longtable}
}

\begin{figure}[H]
\centering
\includegraphics[width=1\textwidth]{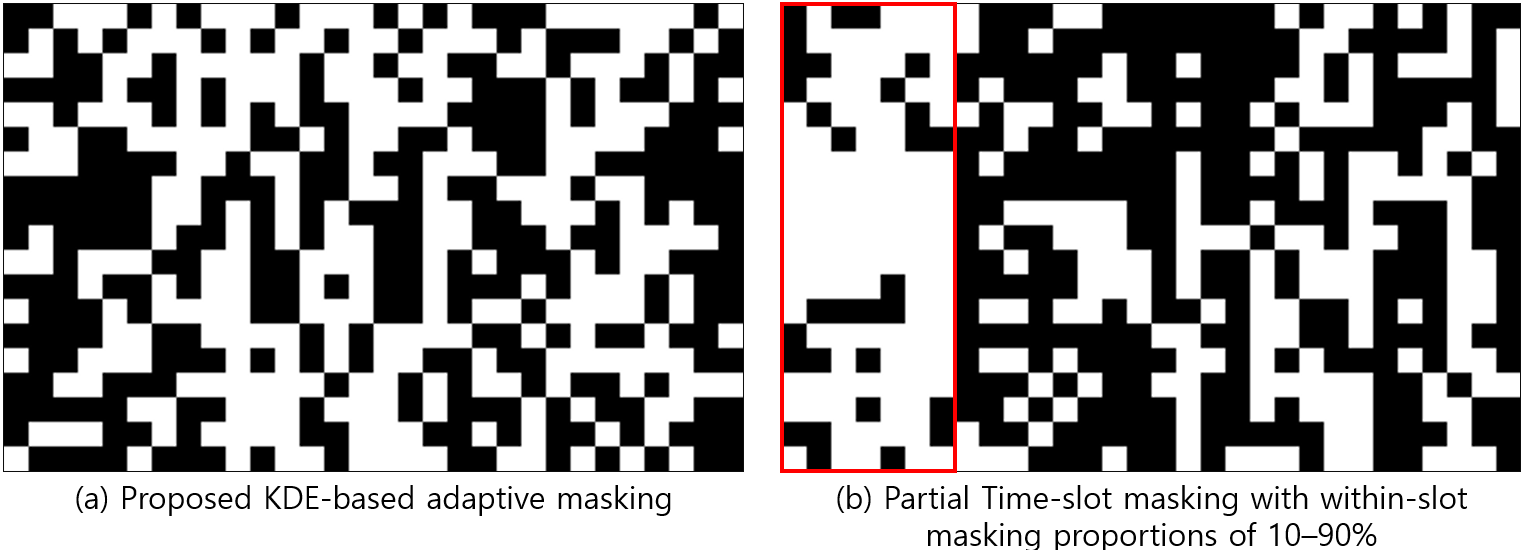}
\caption{Comparison of masking patterns generated by different masking strategies. White cells correspond to masked patches.}\label{fig13} 
\end{figure}

The proposed KDE-based adaptive masking strategy achieved the highest average balanced accuracy (66.03\%) across the 10 datasets, outperforming all alternative masking strategies on average. Although Channel-wise, Time-slot, and Partial Time-slot masking incorporate EEG-specific structures, they were not inferior to the proposed KDE-based masking method but also achieved lower performance than random masking.

Partial Time-slot Masking can increase the occurrence of heavily masked time slots and therefore satisfies the first objective described in Section 3.2, namely reducing the availability of strongly correlated channel information. However, because the time slots are selected independently at random, it does not explicitly prevent simultaneous heavy masking of adjacent time slots. Consequently, it fails to satisfy the second objective described in Section 3.2, namely preserving temporal context by avoiding consecutive heavily masked time slots during reconstruction. These findings suggest that simply increasing the masking density of specific time slots is insufficient for effective representation learning. Additionally, preventing simultaneous heavy masking of adjacent time slots is an important design component contributing to the effectiveness of the proposed KDE-based adaptive masking strategy.

Figure 13 provides a visual comparison between the masking patterns generated by the proposed KDE-based adaptive masking strategy and Partial Time-slot Masking with within-slot masking proportions of 10–90\%. As shown in Figure 13(a), the proposed method distributes heavily masked regions while avoiding long runs of consecutive heavily masked time slots. In contrast, Figure 13(b) shows several adjacent time slots with a high masking density, highlighted by the red box. Because Partial Time-slot Masking selects time slots independently at random, heavily masked time slots can occur consecutively without any mechanism to discourage such patterns. Thus, temporal dependencies across neighboring time slots may become unavailable for reconstruction. This example highlights a key difference between the two strategies: The proposed method preserves temporal context by avoiding consecutive heavily masked time slots, whereas Partial Time-slot Masking can unintentionally remove temporal information required for reconstruction.\\

\textbf{Comparison with Alternative GCN Architectures.}
To justify the use of GCN as a CPE mechanism, we compare the proposed architecture with two alternative GCN-based designs. The first variant, denoted as \textit{GCN-Only}, removes the transformer encoder and performs both pretraining and downstream classification using only GCN features. The second variant, denoted as \textit{GCN-Feature}, uses the same graph structure as the proposed method but incorporates the GCN output as an additional feature rather than as positional encoding. Specifically, the GCN features are added to the output of the transformer encoder before the classification head. In contrast, CABrain(PE) employs GCN outputs as CPEs by adding them to the patch embeddings before transformer processing.

The results are presented in Table 11. Overall, CABrain(PE) achieved the highest average balanced accuracy (65.19\%) among the three architectures, followed by GCN-Feature (63.46\%) and GCN-Only (55.34\%). The results indicate that graph convolution alone is insufficient to fully capture the complex temporal and spatial dependencies required for diverse EEG decoding tasks. Nevertheless, GCN-Only produced competitive results on several datasets. For example, its performance on Mumtaz2016 exceeded that of the proposed method, and its performance on SEED-V surpassed that reported for the early EEG foundation model BIOT. Furthermore, on FACED, PhysioNet-MI, ISRUC, and CHB-MIT, GCN-Only achieved higher balanced accuracies than EEGNet. However, because EEGNet is a lightweight task-specific model trained without large-scale pretraining, these comparisons should be interpreted with caution.

{\footnotesize
\begin{longtable}{l ccc}

\caption{Comparison of balanced accuracies of alternative GCN architectures.}
\label{table11}\\

\toprule
Dataset & \multicolumn{3}{c}{GCN Integration Strategy} \\
\cmidrule(lr){2-4}
 & GCN-only & GCN-Feature & CABrain(PE) \\
\midrule
\endfirsthead

\caption[]{Comparison of balanced accuracies of alternative GCN architectures. (continued)} \\
\toprule
Dataset & \multicolumn{3}{c}{GCN Integration Strategy} \\
\cmidrule(lr){2-4}
 & GCN-only & GCN-Feature & CABrain(PE) \\
\midrule
\endhead

\bottomrule
\endfoot

\bottomrule
\endlastfoot

MentalArithmetic & 56.11 & 73.40 & \textbf{73.47} \\
Mumtaz2016       & \textbf{90.55} & 90.36 & 88.72 \\
SHU-MI           & 56.40 & 61.67 & \textbf{62.82} \\
PhysioNet-MI     & 58.22 & 62.05 & \textbf{63.26} \\
BCIC-IV-2a       & 44.48 & 53.68 & \textbf{55.40} \\
CHB-MIT          & 63.04 & 70.01 & \textbf{74.62} \\
FACED            & 42.92 & 55.78 & \textbf{58.08} \\
SEED-V           & 38.55 & 40.11 & \textbf{40.57} \\
ISRUC            & 73.54 & 78.46 & \textbf{79.19} \\
BCIC2020-3       & 29.63 & 49.12 & \textbf{55.79} \\

\midrule
Average          & 55.34 & 63.46 & \textbf{65.19} \\
\bottomrule

\end{longtable}
}

GCN-Feature achieved an average balanced accuracy of 63.46\%, outperforming GCN-Only (55.34\%) but falling short of the proposed CABrain(PE) architecture (65.19\%) by 1.73 percentage points. These results suggest that the effectiveness of GCN is determined not only by the graph representation itself but also by how it is integrated into the model. Simply adding GCN outputs to the transformer encoder output provides limited benefits, whereas incorporating GCN outputs into the patch embeddings as conditional positional information yields superior average performance. This indicates that correlation-aware channel relationships can be more effectively exploited when they guide the self-attention process throughout the transformer encoder. This observation supports the design choice of implementing GCN as a CPE module rather than as a standalone backbone or a post-encoder feature fusion component.

\subsection{Pretraining Visualization}
\label{subsec4-5}

\begin{figure}[H]
\centering
\includegraphics[width=0.95\textwidth]{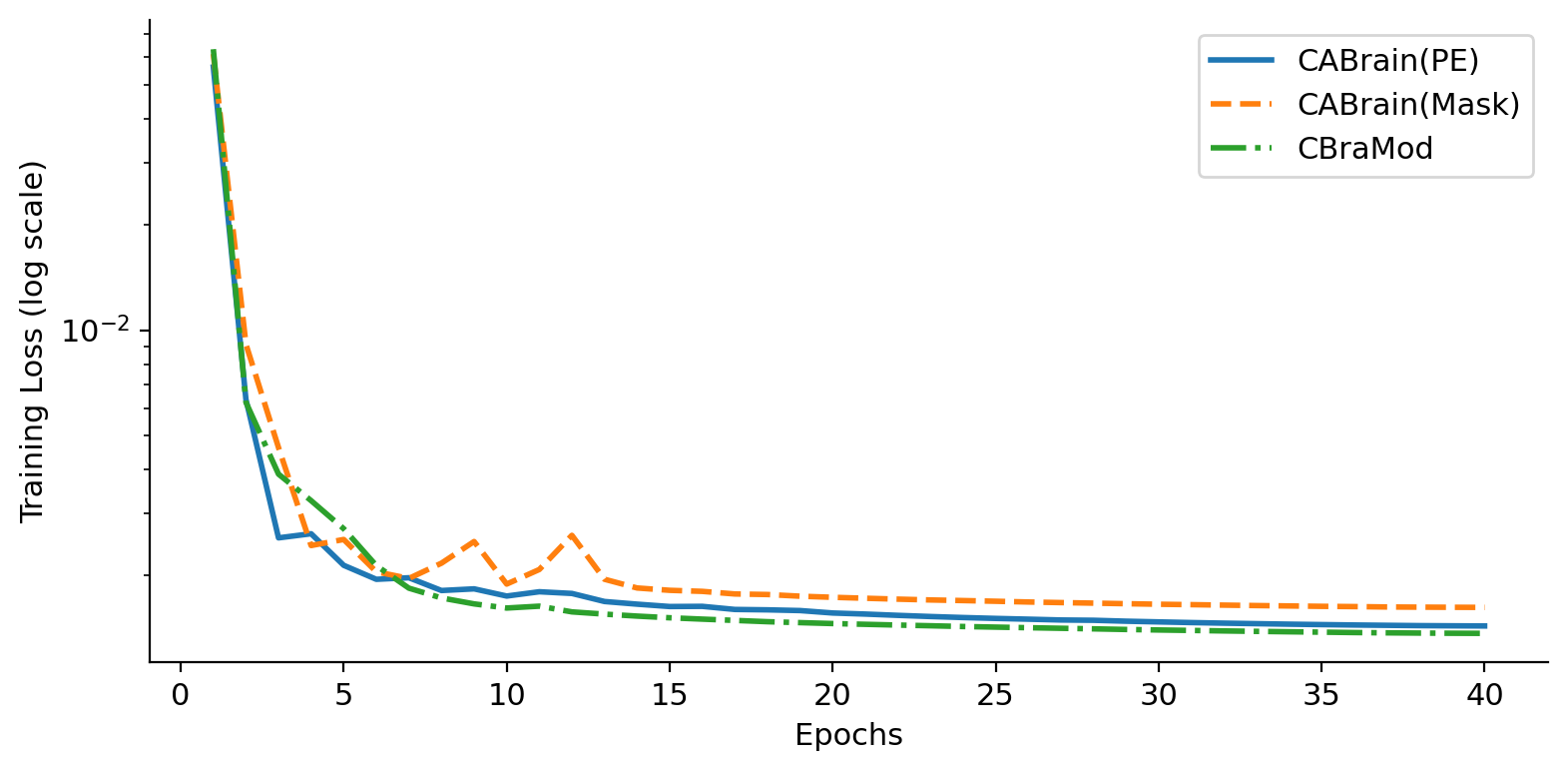}
\caption{Pretraining loss curve.}\label{fig14}
\end{figure}

\begin{figure}[H]
\centering
\includegraphics[width=1\textwidth]{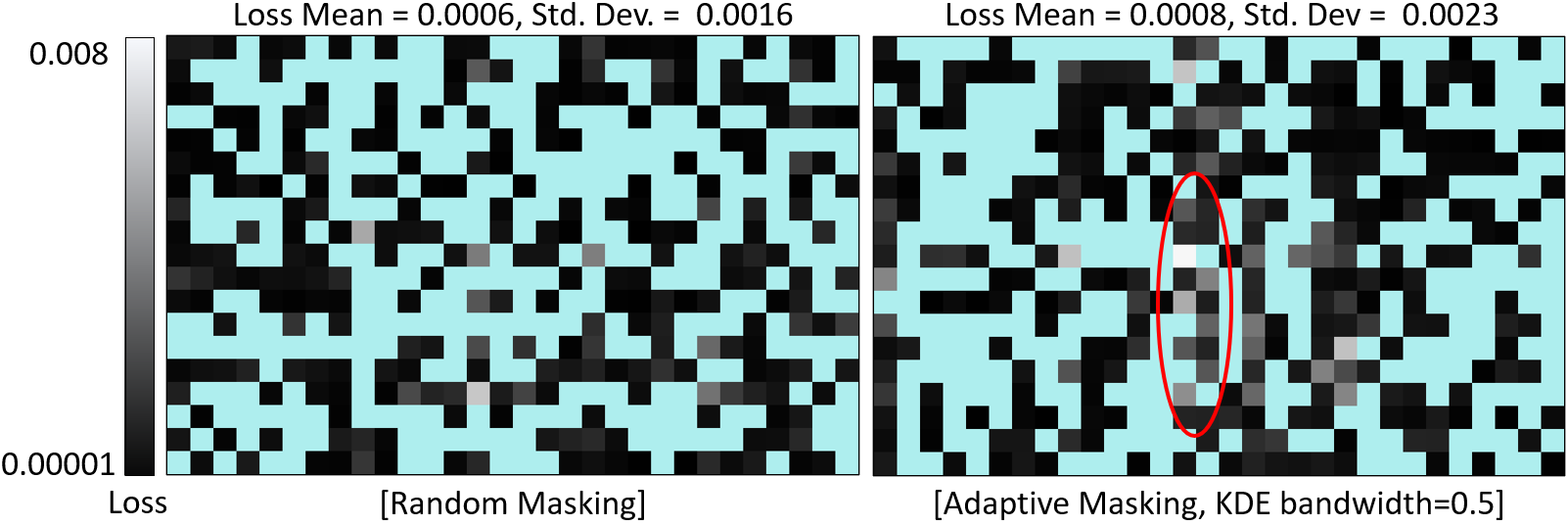}
\caption{Comparison of patch-level loss depending on masking scheme.}\label{fig15}
\end{figure}

Figure 14 presents the pretraining loss curves of CABrain and the previous best-performing baseline CBraMod \cite{CBraMod}. All models converged by epoch 40, with CBraMod achieving the lowest final training loss. This indicates that the baseline model is easier to optimize during pretraining. However, when evaluated on multiple test datasets, both CABrain(PE) and CABrain(Mask) frequently outperformed CBraMod. These findings suggest that, although CBraMod achieves a marginally lower training loss, the proposed methods provide more transferable and generalizable representations. In particular, adaptive masking (CABrain(Mask)) appears to encourage the learning of more robust feature representations, resulting in improved downstream performance despite a slightly higher pretraining loss.

The adaptive masking scheme reduces the contribution of strongly correlated channels and shifts the focus toward weaker or serial correlations. Consequently, the information available for reconstructing masked EEG patches becomes less reliable, which increases the pretraining loss. This trend is observed in Figure 14, with CABrain(Mask) attaining the highest final loss. Increased loss is also evident at the patch level. In Figure 15, the gray-scale intensity represents masked patches, with brighter tones indicating higher reconstruction loss. When adaptive masking was applied (right panel), a larger number of patches exhibited high losses, particularly in the regions highlighted by red circles. In these time slots, the masking ratio was high, and the lack of strongly or moderately correlated channel information resulted in increased reconstruction loss.

\subsection{Representation visualization}
\label{subsec4-6}

To evaluate the discriminative power of the feature representations learned by CABrain, visualizations were obtained using UMAP \cite{UMAP}. Specifically, we compared the visualizations of the raw EEG signals with those of the encoded features after fine-tuning the pretrained CABrain model on the test dataset. 

\begin{figure}[H]
\centering
\includegraphics[width=1\textwidth]{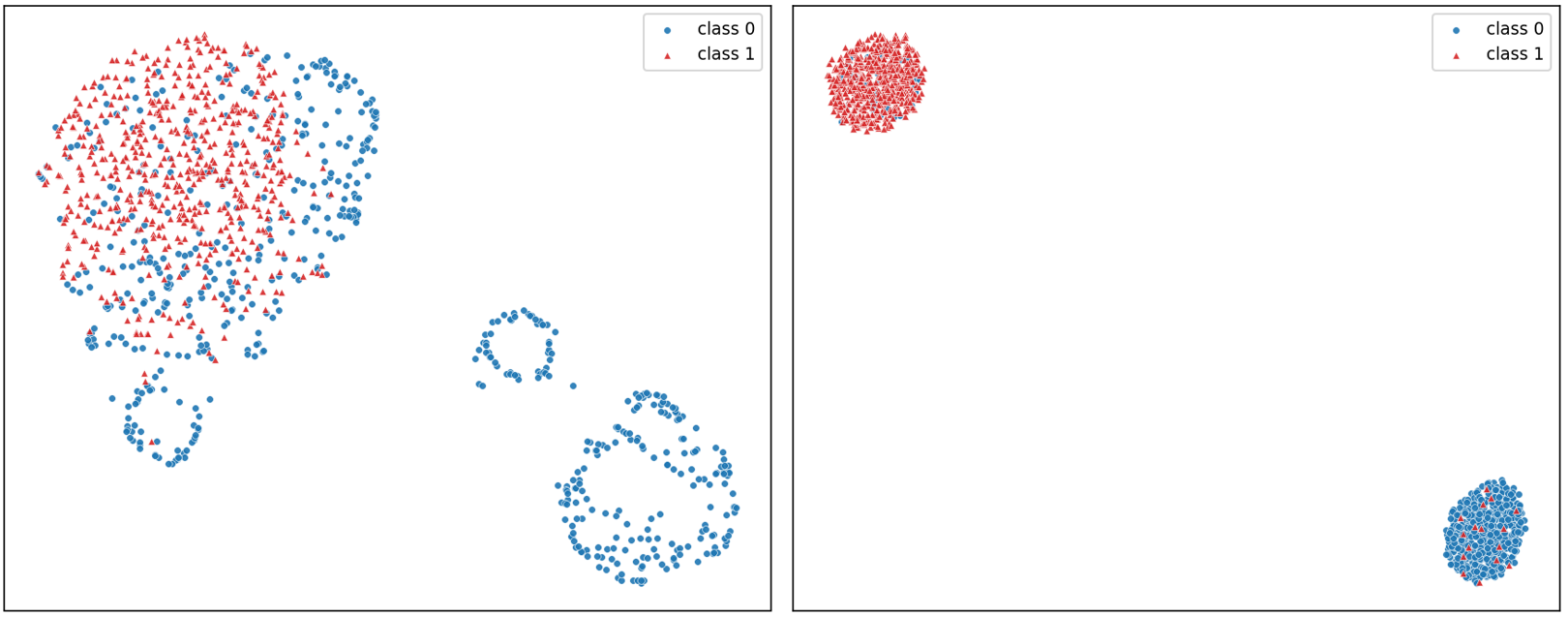}
\caption{UMAP visualization of feature representations on Mumtaz2016. Left: raw EEG signals; Right: CABrain representations.}\label{fig16}
\end{figure}

\begin{figure}[H]
\centering
\includegraphics[width=1\textwidth]{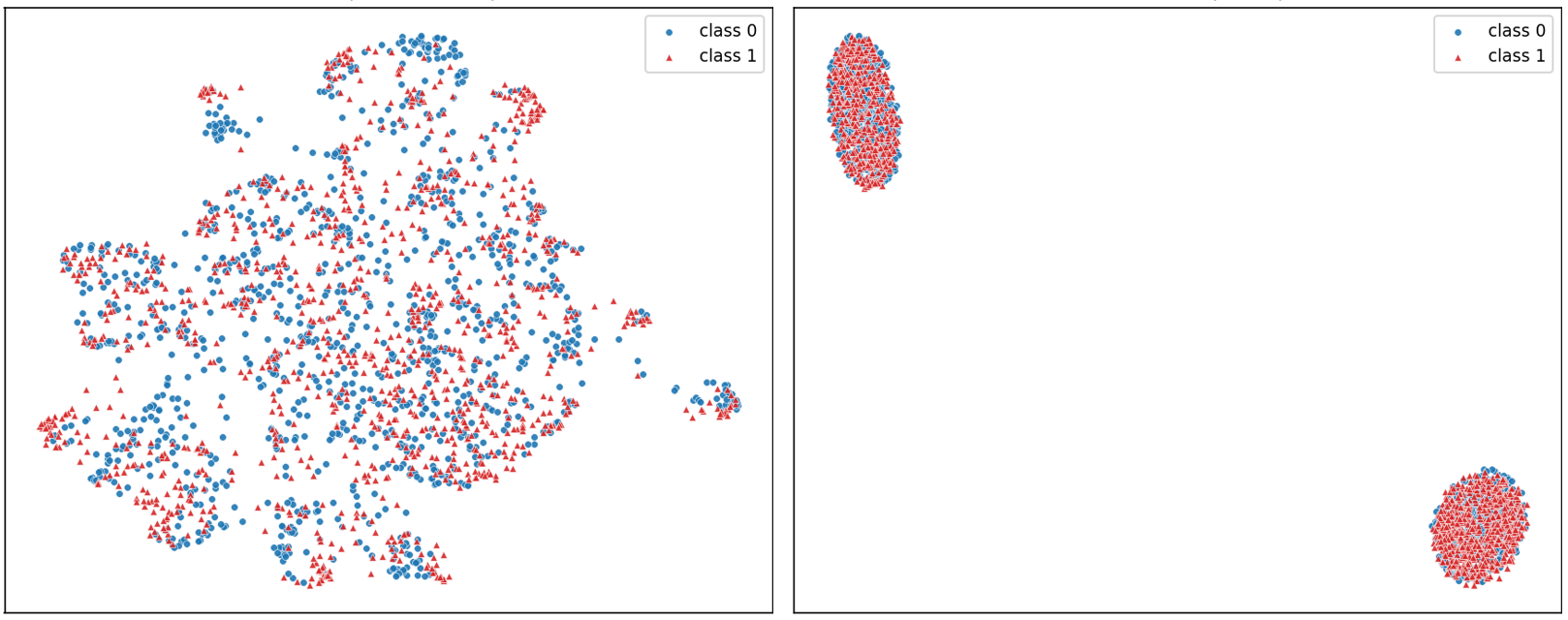}
\caption{UMAP visualization of feature representations on SHU-MI. Left: raw EEG signals; Right: CABrain representations.}\label{fig17}
\end{figure}

\begin{figure}[H]
\centering
\includegraphics[width=1\textwidth]{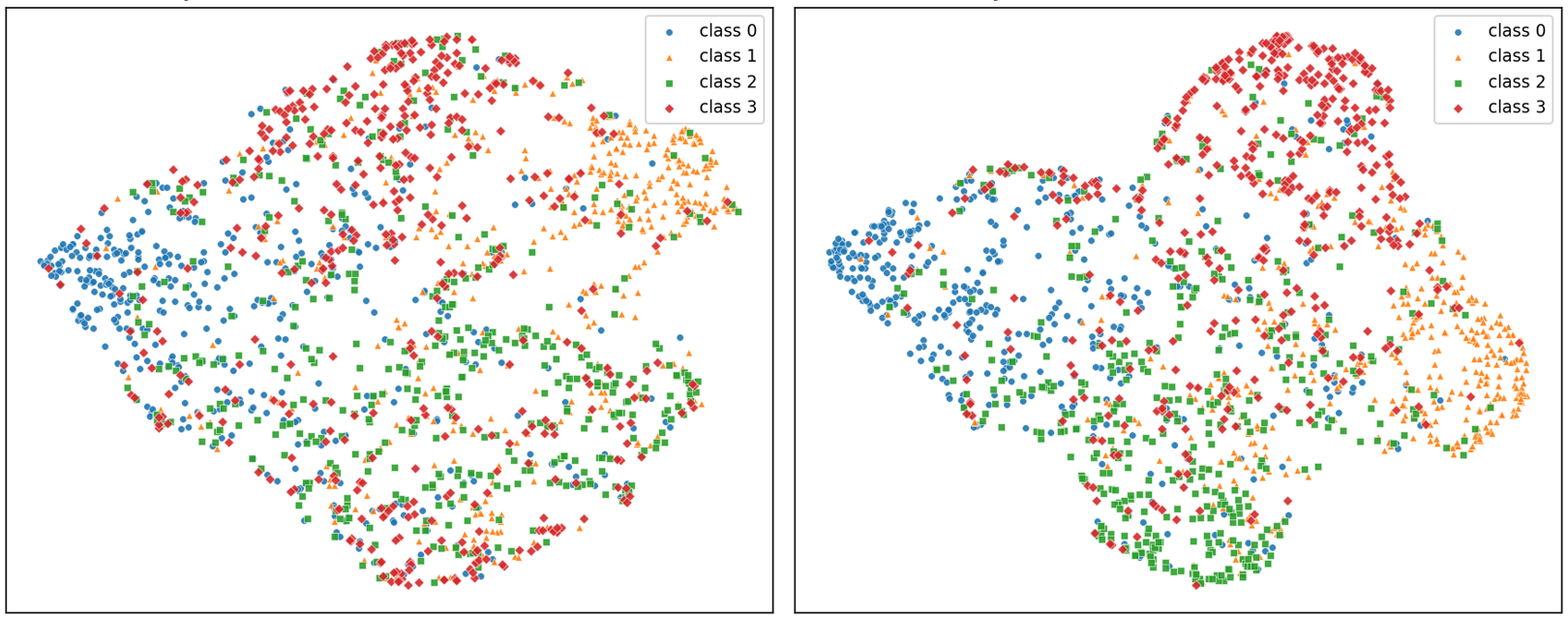}
\caption{UMAP visualization of feature representations on PhysioNet-MI. Left: raw EEG signals; Right: CABrain representations.}\label{fig18}
\end{figure}

Figures 16 and 17 illustrate the visualization results for the Mumtaz2016 \cite{Mumtaz} and SHU-MI \cite{SHU-MI} datasets, respectively, consisting of two classes. While the raw EEG signals projected into two dimensions showed intermingled data points from different classes, the encoded features from the CABrain model formed well-separated clusters based on class. Moreover, the classification accuracy of the Mumtaz2016 dataset was higher than that of the SHU-MI dataset, consistent with the observation that fewer data points from other classes were mixed within the clusters in the Mumtaz2016 dataset. 

Figure 18 presents the visualization results for the PhysioNet-MI \cite{PhySioNet} dataset, which contains four classes. Although the separability of the clusters weakened as the number of classes increased, the encoded features from the CABrain model exhibited stronger class discriminability than raw EEG signals.

\section{Conclusion}
\label{sec5}

This paper introduces CABrain, a correlation-aware EEG foundation model that integrates GCN–based CPE and an adaptive masking strategy for self-supervised pretraining. Extensive experiments on 10 diverse test datasets demonstrate that each approach individually achieves performance comparable or superior to that of state-of-the-art methods, while their combination often yields further improvements. Notably, both approaches are designed following an in-depth analysis of the key factors limiting the performance gains of existing methods. This highlights the systematic nature of our work and its contribution to overcoming fundamental challenges in EEG foundation model pretraining.

Beyond addressing the limitations of existing methods, this study offers a novel perspective rather than incremental refinements. Specifically, this study represents the first attempt at incorporating graph-based representations into CPE, which previously relied on 2D grid inputs. In addition, we propose an adaptive masking strategy for SSL, which has been extensively studied in the computer vision and natural language processing domains but, to the best of our knowledge, has not been previously applied to EEG foundation models. Both proposed methods leverage the correlation information inherent in EEG signals, thereby reflecting a unified design principle. As correlation is not an absolute measure---unlike the number or arrangement of electrodes or the amplitude of signals---it is less sensitive to dataset variability. 

A limitation of the proposed method is the difficulty of directly integrating the two proposed approaches. Because adaptive masking substantially alters the node features provided to the GCN-based CPE, we adopt a transfer-learning strategy instead of applying them simultaneously. Although this indirect integration has been empirically validated, future research should explore more principled fusion strategies to fully leverage their complementary strengths and achieve further performance improvements.

\section*{CRediT authorship contribution statement}
\textbf{SuGil Choi}: Writing – original draft, Writing – review and editing, Conceptualization, Software, Methodology, Formal analysis, Investigation, Visualization. \textbf{Mooseop Kim}: Writing – original draft, Writing – review and editing, Conceptualization, Software, Methodology, Formal analysis, Investigation, Visualization. \textbf{Chi Yoon Jeong}: Writing – review and editing, Validation, Supervision, Project administration, Formal analysis, Funding acquisition, Conceptualization.

\section*{Data availability}
The datasets used in this study are publicly available online.
 
\section*{ACKNOWLEDGMENT}
This work was supported by the Electronics and Telecommunications Research Institute (ETRI) grants funded by the Korean government [26ZR1200, Research on Autonomous Vision Augmentation and Extension Technologies].


\begin{thebibliography}{00}

\bibitem{Boonyakitanont2020}
Poomipat Boonyakitanont, Apiwat Lek-Uthai, Krisnachai Chomtho, and Jitkomut Songsiri. A review of feature extraction and performance evaluation in epileptic seizure detection using EEG. Biomedical Signal Processing and Control, 57, 2020.
\bibitem{Aboalayon2016}
Khald Ali I. Aboalayon, Miad Faezipour, Wafaa S. Almuhammadi, and Saeid Moslehpour. Sleep Stage Classification Using EEG Signal Analysis: A Comprehensive Survey and New Investigation. Entropy, 18(9), 2016.
\bibitem{Jiquan2024}
Jiquan Wang, Sha Zhao, Haiteng Jiang, Shijian Li, Tao Li, and Gang Pan. Generalizable sleep staging via multi-level domain alignment. AAAI Conference on Artificial Intelligence, 2024.
\bibitem{PhySioNet}
Gerwin Schalk, Dennis J McFarland, Thilo Hinterberger, Niels Birbaumer, and Jonathan R Wolpaw. BCI2000: a general-purpose brain-computer interface (BCI) system. IEEE Transactions on biomedical engineering, 51(6), 2004.
\bibitem{SHU-MI}
Jun Ma, Banghua Yang, Wenzheng Qiu, Yunzhe Li, Shouwei Gao, and Xinxing Xia. A large EEG dataset for studying cross-session variability in motor imagery brain-computer interface. Scientific Data, 9(1), 2022.
\bibitem{BCIC2020}
Ji-Hoon Jeong, Jeong-Hyun Cho, Young-Eun Lee, Seo-Hyun Lee, Gi-Hwan Shin, Young-Seok Kweon, Jos´e del R Mill´an, Klaus-Robert M¨uller, and Seong-Whan Lee. 2020 international brain–computer interface competition: A review. Frontiers in Human Neuroscience, 16, 2022.
\bibitem{SEED-VIG}
Jianliang Min, Ping Wang, and Jianfeng Hu. Driver fatigue detection through multiple entropy fusion analysis in an EEG-based system.  PLoS ONE 12(12), 2017.
\bibitem{MentalArithmetic}
Igor Zyma, Sergii Tukaev, Ivan Seleznov, Ken Kiyono, Anton Popov, Mariia Chernykh, and Oleksii Shpenkov. Electroencephalograms during Mental Arithmetic Task Performance. Data, 4(1), 2019.
\bibitem{Sharma}
Lakhan Dev Sharma, Vijay Kumar Bohat, Maria Habib, Ala’ M. Al-Zoubi, Hossam Faris, and Ibrahim Aljarah. Evolutionary inspired approach for mental stress detection using EEG signal. Expert Systems with Applications, 197, 2022.
\bibitem{SEED-V}
Wei Liu, Jie-Lin Qiu, Wei-Long Zheng, and Bao-Liang Lu. Comparing Recognition Performance and Robustness of Multimodal Deep Learning Models for Multimodal Emotion Recognition. IEEE Transactions on Cognitive and Developmental Systems, 14(2), 2021.
\bibitem{FACED}
Jingjing Chen, Xiaobin Wang, Chen Huang, Xin Hu, Xinke Shen, and Dan Zhang. A Large Finer-grained Affective Computing EEG Dataset. Scientific Data, 10(1), 2023.
\bibitem{Bashashati}
Ali Bashashati, Mehrdad Fatourechi, Rabab K Ward and Gary E Birch. A survey of signal processing algorithms in brain–computer interfaces based on electrical brain signals. Journal of Neural Engineering, 4, 2007.
\bibitem{Yudhajit}
Yudhajit Das, Hanli Liu, Fenghua Tian, Srinivas Kota, Rong Zhang, and Lina F. Chalak. Rigor of Neurovascular Coupling (NVC) Assessment in Newborns Using Different Amplitude EEG Algorithms. Scientific Reports, 10, 2020.
\bibitem{EEGNet}
Vernon J Lawhern, Amelia J Solon, Nicholas R Waytowich, Stephen M Gordon, Chou P Hung, and Brent J Lance. EEGNet: a compact convolutional neural network for EEG-based brain–computer interfaces. Journal of Neural Engineering, 15, 2018.
\bibitem{SPaRCNet}
Jin Jing, Wendong Ge, Shenda Hong, Marta Bento Fernandes, Zhen Lin, Chaoqi Yang, Sungtae An, Aaron F Struck, Aline Herlopian, Ioannis Karakis, et al. Development of Expert-Level Classification of Seizures and Rhythmic and Periodic Patterns During EEG Interpretation. Neurology, 100, 2023.
\bibitem{Tengfei}
Tengfei Song, Wenming Zheng, Peng Song, and Zhen Cui. EEG Emotion Recognition Using Dynamical Graph Convolutional Neural Networks. IEEE Transactions on Affective Computing, 11, 2018.
\bibitem{Gaowei}
Gaowei Xu, Tianhe Ren, Yu Chen, and Wenliang Che. A One-Dimensional CNN-LSTM Model for Epileptic Seizure Recognition Using EEG Signal Analysis. Frontiers in Neuroscience, 14, 2020.
\bibitem{Hongli}
Hongli Li, Man Ding, Ronghua Zhang, and Chunbo Xiu. Motor imagery EEG classification algorithm based on CNN-LSTM feature fusion network. Biomedical signal processing and control, 72, 2022.
\bibitem{Song}
Yonghao Song, Qingqing Zheng, Bingchuan Liu, and Xiaorong Gao. EEG Conformer: Convolutional Transformer for EEG Decoding and Visualization. IEEE Transactions on Neural Systems and Rehabilitation Engineering, 31, 2022.
\bibitem{CNN-Transformer}
Wei Yan Peh, Yuanyuan Yao, and Justin Dauwels. Transformer Convolutional Neural Networks for Automated Artifact Detection in Scalp EEG. International Conference of the IEEE Engineering in Medicine and Biology Society (EMBC), 2022.
\bibitem{YilinWang}
Yilin Wang, Sha Zhao, Haiteng Jiang, Shijian Li, Benyan Luo, Tao Li, and Gang Pan. DiffMDD: A Diffusion-Based Deep Learning Framework for MDD Diagnosis Using EEG. IEEE Transactions on Neural Systems and Rehabilitation Engineering, 32, 2024.
\bibitem{BENDR}
Demetres Kostas, Stephane Aroca-Ouellette, and Frank Rudzicz. Bendr: Using transformers and a contrastive self-supervised learning task to learn from massive amounts of eeg data. Frontiers in Human Neuroscience, 15, 2021.
\bibitem{BIOT}
Chaoqi Yang, M Westover, and Jimeng Sun. BIOT: Biosignal Transformer for Cross-data Learning in the Wild. Neural Information Processing Systems, 2023.
\bibitem{Neuro-GPT}
Wenhui Cui, Woojae Jeong, Philipp Thölke, Takfarinas Medani, Karim Jerbi, Anand A. Joshi, and Richard M. Leahy. Neuro-GPT: Towards A Foundation Model For EEG. IEEE International Symposium on Biomedical Imaging, 2024.
\bibitem{NeuroLM}
Wei-Bang Jiang, Yansen Wang, Bao-Liang Lu, and Dongsheng Li. NeuroLM: A Universal Multi-task Foundation Model for Bridging the Gap between Language and EEG Signals. International Conference on Learning Representations, 2025.
\bibitem{BrainGPT}
Tongtian Yue, Xuange Gao, Shuning Xue, Yepeng Tang, Longteng Guo, Jie Jiang, and Jing Liu. BrainGPT: Unleashing the Potential of EEG Generalist Foundation Model by Autoregressive pre-training. arXiv preprint arXiv:2410.19779, 2025.
\bibitem{THD-BAR}
Wenchao Yang, Weidong Yan, Wenkang Liu, Yulan Ma, and Yang Li. THD-BAR: Topology Hierarchical Derived Brain Autoregressive Modeling for EEG Generic Representations. Neural Information Processing Systems, 2025.
\bibitem{FoME}
Enze Shi, Kui Zhao, Qilong Yuan, Jiaqi Wang, Huawen Hu, Sigang Yu, and Shu Zhang. FoME: A Foundation Model for EEG using Adaptive Temporal-Lateral Attention Scaling. arXiv preprint arXiv:2409.12454, 2024.
\bibitem{CBraMod}
Jiquan Wang, Sha Zhao, Zhiling Luo, Yangxuan Zhou, Haiteng Jiang, Shijian Li, Tao Li, and Gang Pan. CBraMod: A Criss-Cross Brain Foundation Model for EEG Decoding. International Conference on Learning Representations, 2025.
\bibitem{EEGMamba}
Jiquan Wang, Sha Zhao, Zhiling Luo, Yangxuan Zhou, Shijian Li, and Gang Pan. EEGMamba: An EEG foundation model with Mamba. Neural Networks, 192, 2025.
\bibitem{CSBrain}
Yuchen Zhou, Jiamin Wu, Zichen Ren, Zhouheng Yao, Weiheng Lu, Kunyu Peng, Qihao Zheng, Chunfeng Song, Wanli Ouyang, and Chao Gou. CSBrain: A Cross-scale Spatiotemporal Brain Foundation Model for EEG Decoding. Neural Information Processing Systems, 2025.
\bibitem{REVE}
Yassine El Ouahidi, Jonathan Lys, Philipp Thölke, Nicolas Farrugia, Bastien Pasdeloup, Vincent Gripon, Karim Jerbi, and Giulia Lioi. REVE: A Foundation Model for EEG - Adapting to Any Setup with Large-Scale Pretraining on 25,000 Subjects. Neural Information Processing Systems, 2025.
\bibitem{LUNA}
Berkay Döner, Thorir Mar Ingolfsson, Luca Benini, and Yawei Li. LUNA: Efficient and Topology-Agnostic Foundation Model for EEG Signal Analysis. Neural Information Processing Systems, 2025.
\bibitem{EEGPT}
Guangyu Wang, Wenchao Liu, Yuhong He, Cong Xu, Lin Ma, and Haifeng Li. EEGPT: Pretrained Transformer for Universal and Reliable Representation of EEG Signals. Neural Information Processing Systems, 2024.
\bibitem{EEG2Rep}
Navid Mohammadi Foumani, Geoffrey Mackellar, Soheila Ghane, Saad Irtza, Nam Nguyen, and Mahsa Salehi. EEG2Rep: Enhancing Self-supervised EEG Representation Through Informative Masked Inputs. ACM SIGKDD Conference on Knowledge Discovery and Data Mining, 2024.
\bibitem{MIRepNet}
Dingkun Liu, Zhu Chen, Jingwei Luo, Shijie Lian, Yuheng Chen, Shaojie Hou, Xiaolian Zhu, and Dongrui Wu. MIRepNet: A pipeline and pre-trained model for EEG-based motor imagery classification. Knowledge-Based Systems, 2026.
\bibitem{LaBraM}
Weibang Jiang, Liming Zhao, and Bao liang Lu. Large Brain Model for Learning Generic Representations with Tremendous EEG Data in BCI. International Conference on Learning Representations, 2024.
\bibitem{BrainOmni}
Qinfan Xiao, Ziyun Cui, Chi Zhang, SiQi Chen, Wen Wu, Andrew Thwaites, Alexandra Woolgar, Bowen Zhou, and Chao Zhang. BrainOmni: A Brain Foundation Model for Unified EEG and MEG Signals. Neural Information Processing Systems, 2025.
\bibitem{EpilepsyFM}
Zhuoyi Li, Ning Zhu, Yifan Chen, Beibei Chen, Qiufeng Dong, Lin Gan, Shijie Zhao, Zhiqiang Yan, and Tuo Zhang.  EpilepsyFM: A domain-specific foundation model for epileptic representation learning using EEG signals. Neural Networks, 2026.
\bibitem{CodeBrain}
Jingying Ma, Feng Wu, Qika Lin, Yucheng Xing, Chenyu Liu, Ziyu Jia, and Mengling Feng. CodeBrain: Bridging Decoupled Tokenizer and Multi-Scale Architecture for EEG Foundation Model. International Conference on Learning Representations, 2026.
\bibitem{BrainBERT}
Christopher Wang, Vighnesh Subramaniam, Adam Uri Yaari, Gabriel Kreiman, Boris Katz, Ignacio Cases, and Andrei Barbu. BrainBERT: Self-supervised representation learning for intracranial recordings. International Conference on Learning Representations, 2023.
\bibitem{EEGFormer}
Zhijiang Wan, Manyu Li, Shichang Liu, Jiajin Huang, Hai Tan, and Wenfeng Duan. EEGformer: A transformer–based brain activity classification method using EEG signal. Frontiers in Neuroscience, 2023.
\bibitem{BrainWave}
Zhizhang Yuan, Fanqi Shen, Meng Li, Yuguo Yu, Chenhao Tan, and Yang Yang. BrainWave: A Brain Signal Foundation Model for Clinical Applications. arXiv preprint arXiv:2402.10251, 2024.
\bibitem{Uni-NTFM}
Zhisheng Chen, Yingwei Zhang, Qizhen Lan, Tianyu Liu, Huacan Wang, Yi Ding, Ziyu Jia, Ronghao Chen, Kun Wang, and Xinliang Zhou. Uni-NTFM: A Unified Foundation Model for EEG Signal Representation Learning. International Conference on Learning Representations, 2026.
\bibitem{CPE}
Xiangxiang Chu, Zhi Tian, Bo Zhang, Xinlong Wang, and Chunhua Shen. Conditional Positional Encodings for Vision Transformers. International Conference on Learning Representations, 2023.
\bibitem{TUEG}
Iyad Obeid and Joseph Picone. The temple university hospital eeg data corpus. Frontiers in neuroscience, 10, 2016.
\bibitem{RoPE}
Jianlin Su, Murtadha Ahmed, Yu Lu, Shengfeng Pan, Wen Bo, and Yunfeng Liu. Roformer: Enhanced transformer with rotary position embedding. Neurocomputing, 568, 2024.
\bibitem{Kohei}
Kohei Saijo and Tetsuji Ogawa. A Comparative Study on Positional Encoding for Time-frequency Domain Dual-path Transformer-based Source Separation Models. arXiv preprint arXiv:2504.19605, 2025.
\bibitem{BERT}
Jacob Devlin, Ming-Wei Chang, Kenton Lee, and Kristina Toutanova.  Bert: pretraining of deep bidirectional  transformers for language understanding. Conference of the North American Chapter of the Association for Computational Linguistics: Human Language Technologies, 1, 2019.
\bibitem{MASS}
Kaitao Song, Xu Tan, Tao Qin, Jianfeng Lu, Tie-Yan Liu. MASS: Masked Sequence to Sequence pre-training for Language Generation. International Conference on Machine Learning, 2019.
\bibitem{BEIT}
Hangbo Bao, Li Dong, and Furu Wei. Beit: Bert pre-training of image transformers. International Conference on Learning Representations, 2022.
\bibitem{KaimingHe}
Kaiming He, Xinlei Chen, Saining Xie, Yanghao Li, Piotr Doll´ar, and Ross Girshick. Masked autoencoders are scalable vision learners. Conference on Computer Vision and Pattern Recognition, 2022. 
\bibitem{Simmim}
Zhenda Xie, Zheng Zhang, Yue Cao, Yutong Lin, Jianmin Bao, Zhuliang Yao, Qi Dai, and Han Hu. Simmim: A simple framework for masked image modeling. Conference on Computer Vision and Pattern Recognition, 2022. 
\bibitem{APOLLO}
Soumya Sanyal, Yichong Xu, Shuohang Wang, Ziyi Yang, Reid Pryzant, Wenhao Yu, Chenguang Zhu, and Xiang Ren. APOLLO: A Simple Approach for Adaptive Pretraining of Language Models for Logical Reasoning. Annual Meeting of the Association for Computational Linguistics, 1, 2023.
\bibitem{Vanshpreet}
Vanshpreet S. Kohli, Aaron Monis, and Radhika Mamidi. Choose Your Words Wisely: Domain-adaptive Masking Makes Language Models Learn Faster. Workshop on Representation Learning for NLP, 2025. 
\bibitem{ZhanzhouFeng}
Zhanzhou Feng and Shiliang Zhang. Evolved Part Masking for Self-Supervised Learning. Conference on Computer Vision and Pattern Recognition, 2023. 
\bibitem{AdaMAE}
Wele Gedara Chaminda Bandara, Naman Patel, Ali Gholami, Mehdi Nikkhah, Motilal Agrawal, and Vishal M. Patel. AdaMAE: Adaptive Masking for Efficient Spatiotemporal Learning with Masked Autoencoders. Conference on Computer Vision and Pattern Recognition, 2023. 
\bibitem{CSMAE}
Nisarg A. Shah, Chaminda Bandara, Shameema Sikder, S. Swaroop Vedula, Vishal M. Patel. CSMAE : Cataract Surgical Masked Autoencoder (MAE) Based pre-training. International Symposium on Biomedical Imaging, 2025.
\bibitem{LongxuanDai}
Longxuan Dai, Lei Hu, Liangbing Chen, Chenxi Wang, and Feng Lin. An Image Double Encryption Based on Improved GAN and Hyper Chaotic System. IEEE Access, 12, 2024.
\bibitem{SiyiTang}
Siyi Tang, Jared A. Dunnmon, Khaled Saab, Xuan Zhang, Qianying Huang, Florian Dubost, Daniel L. Rubin, and Christopher Lee-Messer. Self-Supervised Graph Neural Networks for Improved Electroencephalographic Seizure Analysis. International Conference on Learning Representations, 2022.
\bibitem{JinpeiHan}
Jinpei Han, Xiaoxi Wei, and A Aldo Faisal. EEG decoding for datasets with heterogenous electrode configurations using transfer learning graph neural networks. Journal of Neural Engineering, 20, 2023.
\bibitem{EdgeDrop}
Yu Rong, Wenbing Huang, Tingyang Xu, and Junzhou Huang. DropEdge: Towards Deep Graph Convolutional Networks on Node Classification. International Conference on Learning Representations, 2020.
\bibitem{BCIC-IV-2a}
Clemens Brunner, Robert Leeb, Gernot M¨uller-Putz, Alois Schl¨ogl, and Gert Pfurtscheller. BCI Competition 2008 Graz data set A. Graz University of Technology, 16, 2008.
\bibitem{ISRUC}
Sirvan Khalighi, Teresa Sousa, Jos´e Moutinho Santos, and Urbano Nunes. ISRUC-Sleep: A comprehensive public dataset for sleep researchers. Computer methods and programs in biomedicine, 124, 2016.
\bibitem{CHB-MIT}
Ali Hossam Shoeb. Application of machine learning to epileptic seizure onset detection and treatment. PhD thesis, Massachusetts Institute of Technology, 2009.
\bibitem{Mumtaz}
Wajid Mumtaz. MDD Patients and Healthy Controls EEG Data (New). 2016. doi: 10.6084/m9.figshare.4244171.v2.
\bibitem{UMAP}
Leland McInnes, John Healy, and James Melville. UMAP: Uniform Manifold Approximation and Projection for Dimension Reduction. arXiv preprint arXiv:1802.03426, 2018.
 
\end{thebibliography}
\end{document}